\documentclass[10pt,twocolumn,letterpaper]{article}

\usepackage{cvpr}
\usepackage{times}
\usepackage{epsfig}
\usepackage{graphicx}
\usepackage{amsmath}
\usepackage{amssymb}
\usepackage{subfigure}
\usepackage{multirow}
\usepackage{bm}

\usepackage[breaklinks=true,bookmarks=false]{hyperref}

\newcommand{\tabincell}[2]{\begin{tabular}{@{}#1@{}}#2\end{tabular}}

\cvprfinalcopy 

\def\cvprPaperID{****} 

\ifcvprfinal\fi

\begin{document}

\title{What Deep CNNs Benefit from Global Covariance Pooling: An Optimization Perspective}

\author{Qilong Wang$^1$, Li Zhang$^1$, Banggu Wu$^1$, Dongwei Ren$^1$, Peihua Li$^2$, Wangmeng Zuo$^3$, Qinghua Hu$^{1,}$\thanks{Qinghua Hu is the corresponding author. 
\newline Email: \{qlwang, li\_zhang, huqinghua\}@tju.edu.cn. The work was supported by the National Natural Science Foundation of China (No. 61806140, 61971086, 61925602, U19A2073, 61732011). Qilong Wang was supported by National Postdoctoral Program for Innovative Talents.} \\ 
$^1$ Tianjin Key Lab of Machine Learning, College of Intelligence and Computing, Tianjin University, China\\
$^2$ Dalian University of Technology, China \,\,\,\,\, $^3$ Harbin Institute of Technology, China\\	
}
\maketitle
\thispagestyle{empty}

\begin{abstract}
	Recent works have demonstrated that global covariance pooling (GCP) has the ability to improve performance of deep convolutional neural networks (CNNs) on visual classification task. Despite considerable advance, the reasons on effectiveness of GCP on deep CNNs have not been well studied. In this paper, we make an attempt to understand what deep CNNs benefit from GCP in a viewpoint of optimization. Specifically, we explore the effect of GCP on deep CNNs in terms of the Lipschitzness of optimization loss and the predictiveness of  gradients, and show that GCP can make the optimization landscape more smooth and the gradients more predictive. Furthermore, we discuss the connection between GCP and second-order optimization for deep CNNs. More importantly, above findings can account for several merits of covariance pooling for training deep CNNs that have not been recognized previously or fully explored, including significant acceleration of network convergence (i.e., the networks trained with GCP can support rapid decay of learning rates, achieving favorable performance while significantly reducing number of training epochs), stronger robustness to distorted examples generated by image corruptions and perturbations, and good generalization ability to different vision tasks, e.g., object detection and instance segmentation. We conduct extensive experiments using various deep CNN models on diversified tasks, and the results provide strong support to our findings.      
\end{abstract}

\section{Introduction}

\begin{figure}[t]
	\begin{center}
		\includegraphics[width=0.88\linewidth]{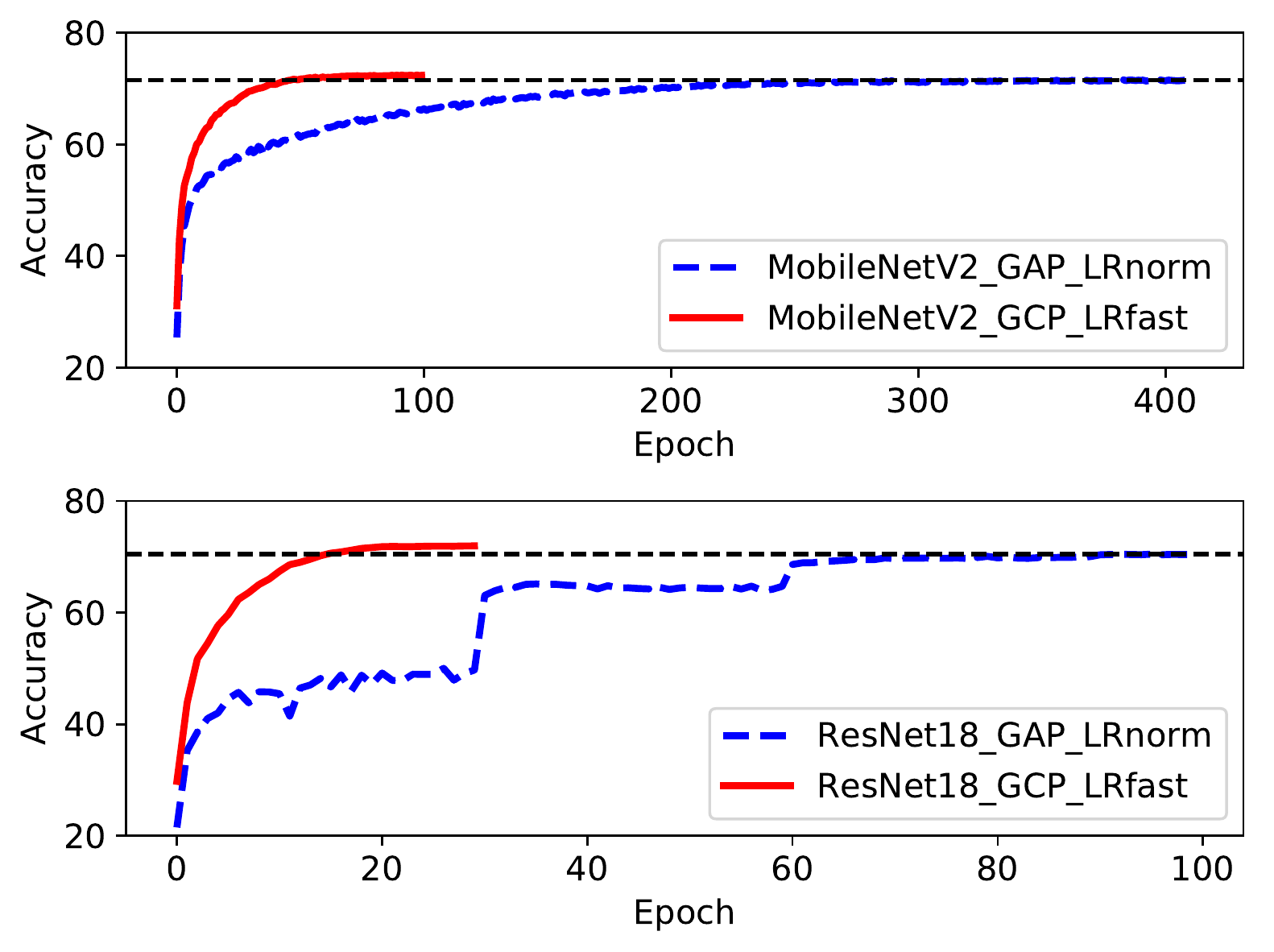}
	\end{center}
	\caption{Convergence curves of MobileNetV2~\cite{DBLP:conf/cvpr/SandlerHZZC18} and ResNet-18~\cite{He_2016_CVPR} with  global average pooling (GAP) and global covariance pooling (GCP) on ImageNet. Note that the networks with GCP converge much faster while achieving matching or better results. We account for it from an optimization perspective (see Section~\ref{main}).}
	\label{fig:motivation}
\end{figure}

Global covariance pooling (GCP) that is used to replace global average pooling (GAP) for aggregating the last convolution activations of deep convolutional neural networks (CNNs) has achieved remarkable performance gains on a variety of vision tasks~\cite{Ionescu_2015_ICCV,lin2015bilinear,DBLP:conf/emnlp/FukuiPYRDR16,Wang_2017_CVPR,DBLP:conf/cvpr/DibaSG17,LiXWZ17,lin2017improved,LiXWG18}. Existing GCP-based works mainly focus on obtaining better performance using various normalization methods~\cite{Ionescu_2015_ICCV,lin2015bilinear,LiXWZ17,lin2017improved} and richer statistics~\cite{Wang_2017_CVPR,Dai_2017_CVPR,Cui_2017_CVPR,Cai_2017_ICCV} or achieving comparable results with low-dimensional covariance representations~\cite{Gao_2016_CVPR,Kong_Charless_2017_CVPR,Gou_2018_CVPR,DBLP:conf/eccv/YuS18}. However, the reasons on effectiveness of GCP on deep CNNs have not been well studied. Although some works explain them from the perspectives of statistical modeling~\cite{Ionescu_2015_ICCV,lin2015bilinear,LiXWZ17} or geometry~\cite{LiXWZ17}, some behaviors of deep CNNs with GCP still lack of reasonable explanations. For example, as illustrated in Figure~\ref{fig:motivation}, why GCP can significantly speed up convergence of deep CNNs. Particularly, the networks with GCP can achieve matching or better performance than GAP-based ones, but only use less than $1/4$ of training epochs of the latter one.  

In this paper, we make an attempt to understand the effectiveness of GCP on deep CNNs from an optimization perspective, thereby explaining behaviors of the networks with GCP in an intuitive way. To this end, we explore the effect of GCP on optimization landscape and gradient computation of deep CNNs, inspired by recent work~\cite{DBLP:conf/nips/SanturkarTIM18}. Specifically, we first train two widely used CNN models (i.e., MobileNetV2~\cite{DBLP:conf/cvpr/SandlerHZZC18} and ResNet-18~\cite{He_2016_CVPR}) on large-scale ImageNet~\cite{imagenet_cvpr09} using GAP and GCP for aggregating the last convolution activations, respectively. Then, we compare them from optimization perspective, and find that GCP can improve the stability of optimization loss (i.e., Lipschitzness) and the stability of gradients (i.e., predictiveness) over the commonly used GAP. Furthermore, by analyzing back-propagation of GCP and second-order optimization~\cite{DBLP:conf/icml/Martens10,DBLP:conf/icml/MartensG15,DBLP:conf/cvpr/OsawaTUNYM19} in the context of deep CNNs, we make out that the influence of GCP on optimization shares some similar philosophy with K-FAC~\cite{DBLP:conf/icml/MartensG15,DBLP:conf/cvpr/OsawaTUNYM19}. Above findings provide an inspiring view to understand effectiveness of GCP on deep CNNs, which can intuitively and reasonably account for the behaviors of deep CNNs trained with GCP, e.g., GCP makes the optimization landscape more smooth, leading to network convergence to a better local minimum (i.e., better performance as shown in~\cite{lin2015bilinear,Wang_2017_CVPR,LiXWZ17}) and much faster convergence as illustrated in Figure~\ref{fig:motivation}.

Based on the foregoing findings, we can explain several merits delivered by GCP for training deep CNNs that have not been recognized previously or fully explored. Firstly, since GCP is able to smoothen optimization landscape, deep CNNs with GCP can support rapid decay of learning rates for fast convergence. Meanwhile, previous work~\cite{DBLP:conf/nips/SanturkarTIM18} shows that improvement of Lipschitzness can accelerate convergence of deep CNNs. To verify this point, we conduct experiments using a variety of deep CNN architectures (i.e., MobileNetV2~\cite{DBLP:conf/cvpr/SandlerHZZC18}, ShuffleNet V2~\cite{DBLP:conf/eccv/MaZZS18} and ResNets~\cite{He_2016_CVPR}) on ImageNet~\cite{imagenet_cvpr09}. The results show that, under the setting of rapid decay of learning rates, deep CNNs with GCP achieve comparable performance to GAP-based ones, while using less than a quarter of training epochs. By adjusting schedule of learning rates, deep CNNs with GCP can converge to better local minima using less training epochs. 

Secondly, GCP improves the stability of both optimization loss and gradients, so it makes deep CNNs more robust to inputs perturbed by some distortions. Meanwhile, previous work~\cite{DBLP:conf/icml/CisseBGDU17} shows that control of the Lipschitz constant is helpful for improving robustness to examples with crafted distortions for confusing classifiers. Therefore, we experiment on recently introduced IMAGENET-C and IMAGENET-P benchmarks~\cite{DBLP:conf/iclr/HendrycksD19}, where distortions are achieved by common image corruptions and perturbations. The results show that GCP can significantly improve robustness of deep CNNs to image corruptions and perturbations. 

Thirdly, GCP usually allows deep CNNs converge to better local minima, and thus deep CNNs pre-trained with GCP can be exploited to provide an effective initialization model for other visual tasks. Therefore, we verify it by transferring the pre-trained CNN models to MS COCO benchmark~\cite{MSCOCO} for object detection and instance segmentation tasks, and the results indicate that pre-trained CNNs  with GCP is superior to GAP-based ones. 

The contributions of this paper are concluded as follows. (1) To our best knowledge, we make the first attempt to understand the effectiveness of GCP in the context of deep CNNs from an optimization perspective. Specifically, we show that GCP can improve the Lipschitzness of optimization loss and the predictiveness of gradients. Furthermore, we discuss the connection between GCP and second-order optimization. These findings provide an inspiring view to better understand the behaviors of deep CNNs trained with GCP. (2) More importantly, our findings above can explain several merits of GCP for training deep CNNs  that have not been recognized previously or fully explored, including significant acceleration of convergence with rapid decay of learning rates, stronger robustness to distorted examples and good generalization ability to different vision tasks. (3) We conduct extensive experiments using six representative deep CNN architectures on image classification, object detection and instance segmentation, the results of which provide strong support to our findings. 

\section{Related Work}

DeepO$_2$P~\cite{Ionescu_2015_ICCV} and B-CNN~\cite{lin2015bilinear} are among the first works introducing GCP into deep CNNs. DeepO$_2$P extends the second-order (covariance) pooling method (O$_2$P)~\cite{carreira_pami14} to deep architecture, while B-CNN captures interactions of localized convolution features by a trainable bilinear pooling. Wang \etal~\cite{Wang_2017_CVPR} propose a global Gaussian distribution embedding network for utilizing the power of probability distribution modeling and deep learning jointly. Li \etal~\cite{LiXWZ17} present matrix power normalization for  GCP and make clear its statistical and geometrical mechanisms. Beyond GCP, some researches~\cite{Cui_2017_CVPR,Cai_2017_ICCV} propose to use richer statistics for further possible improvement. The aforementioned methods study GCP from perspectives of statistical modeling or Riemannian geometry. Different from them, we interpret the effectiveness of GCP from an optimization perspective, and further explore merits of GCP for training deep CNNs.

Since deep CNNs are black-boxes themselves, understanding effect of GCP on deep CNNs still is a challenging issue. Many recent works make  efforts~\cite{ZeilerF14,DBLP:conf/iclr/ZhangBHRV17,DBLP:conf/nips/SanturkarTIM18,DBLP:conf/nips/BjorckGSW18,DBLP:journals/pami/ZhouBO019} towards understanding deep CNNs or analyzing the effect of fundamental components, e.g., individual units, batch normalization (BN)~\cite{icml2015_ioffe15} and optimization algorithm. Specifically, Zeiler \etal~\cite{ZeilerF14} and Zhou \etal~\cite{DBLP:journals/pami/ZhouBO019} visualize feature maps by deconvolution and image regions of maximal activation units, respectively. Zhang \etal~\cite{DBLP:conf/iclr/ZhangBHRV17} and Bjorck \etal~\cite{DBLP:conf/nips/BjorckGSW18} design a series of controlled experiments to analyze generalization ability of deep CNNs and understand effect of BN on deep CNNs, respectively. In particular, a recent work~\cite{DBLP:conf/nips/SanturkarTIM18} investigates the effect of BN by exploring optimization landscape of VGG-like networks~\cite{Simonyan15} trained on CIFAR10~\cite{CIFAR}, while providing theoretical analysis on Lipschitzness improvement using a fully-connected layer. Motivated by~\cite{DBLP:conf/nips/SanturkarTIM18}, we make an attempt to understand \emph{the effect of GCP on deep CNNs} from an optimization perspective.

\section{An Optimization Perspective for GCP}\label{main}

In this section, we first revisit global covariance pooling (GCP) for deep CNNs. Then, we analyze smoothing effect of GCP on deep CNNs, and finally discuss its connection to second-order optimization.   

\subsection{Revisiting  GCP}
Let $\mathcal{X}\in\mathbb{R}^{W\times H\times D}$ be the output of the last convolution layer of deep CNNs, where $W$, $H$ and $D$ indicate width, height and dimension of feature map, respectively. To summarize $\mathcal{X}$ as global representations for final prediction, most existing CNNs employ GAP, i.e., $\sum_{i=1}^{N}\mathbf{X}_{i}$, where the feature tensor $\mathcal{X}$ is reshaped to a feature matrix $\mathbf{X}\in\mathbb{R}^{N \times D}$ and $N=W \times H$. Many recent works~\cite{lin2015bilinear,LiXWZ17} demonstrate the superiority of GCP over GAP. To perform GCP, the sample covariance matrix of $\mathbf{X}$ can be calculated as
\begin{align}\label{COV}
\boldsymbol{\Sigma} = \mathbf{X}^{T}\mathbf{J}\mathbf{X},\,\, \mathbf{J} = \frac{1}{N}(\mathbf{I}-\frac{1}{N}\mathbf{1}\mathbf{1}^{T}), 
\end{align}
where $\mathbf{I}$ is the $N \times N$ identity matrix, and $\mathbf{1}$ is a $N$-dimensional vector of all elements being one. 

Normalization plays an important role in GCP, and different normalization methods have been studied, including matrix logarithm normalization~\cite{Ionescu_2015_ICCV}, element-wise power normalization followed by $\ell_{2}$ normalization~\cite{lin2015bilinear}, matrix square-root normalization~\cite{Wang_2017_CVPR,lin2017improved,LiXWG18} and matrix power normalization~\cite{LiXWZ17}. Among them, matrix square-root (i.e., the power of $1/2$) normalization is preferred considering its promising performance on both large-scale and small-scale visual classification tasks. Therefore, this paper uses GCP with matrix square-root normalization, i.e.,
\begin{align}\label{SR_COV}
\boldsymbol{\Sigma}^{\frac{1}{2}}
=\mathbf{U}\boldsymbol{\Lambda}^{\frac{1}{2}}\mathbf{U}^{T}, 
\end{align} 
where $\mathbf{U}$ and $\boldsymbol{\Lambda}$ are the matrix of eigenvectors and the diagonal matrix of eigenvalues of  $\boldsymbol{\Sigma}$, respectively. 

\begin{figure}[t]
	\begin{center}
		\includegraphics[width=1.0\linewidth]{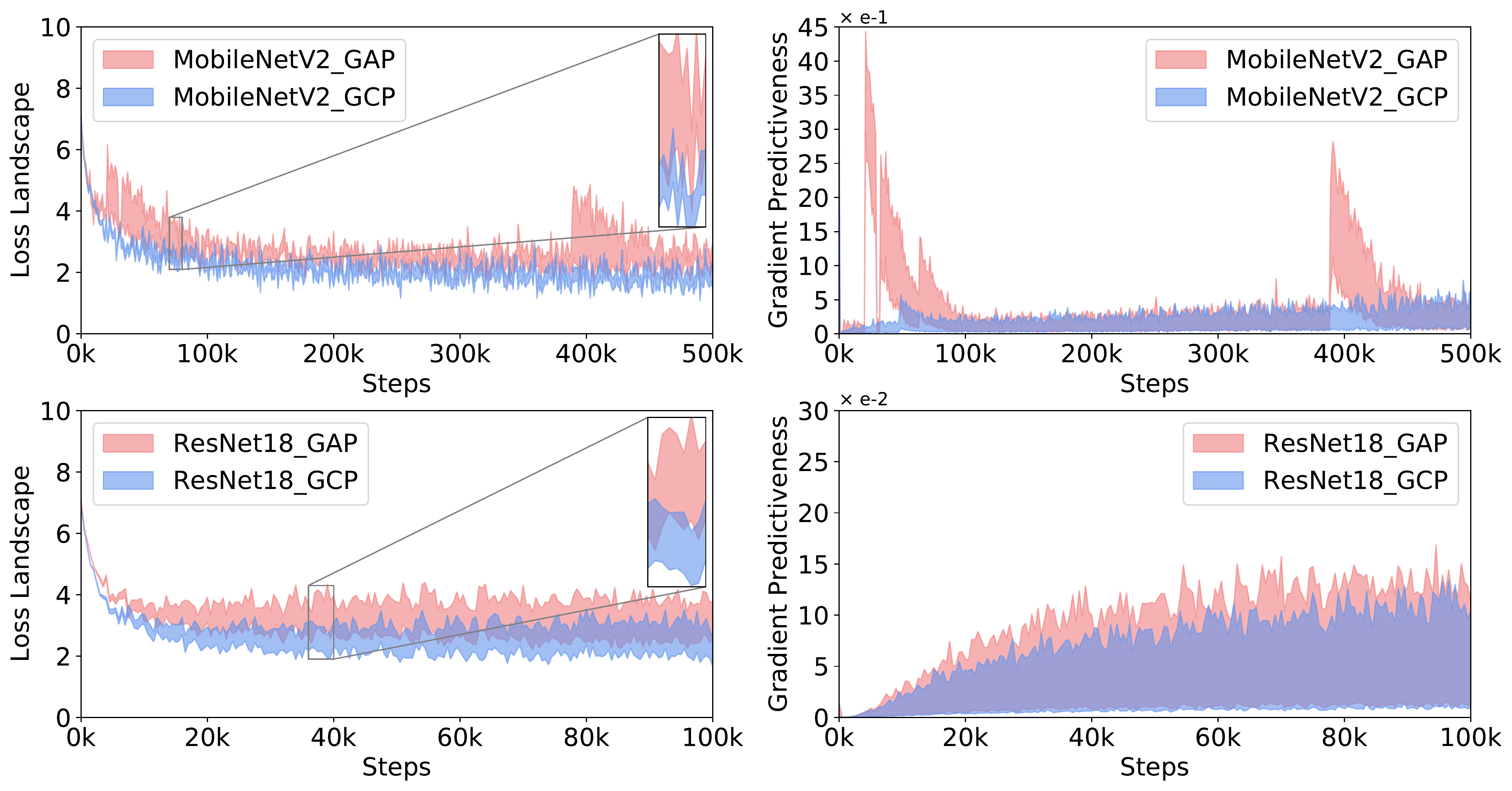}
	\end{center}
	\caption{Comparison of (top) MobileNetV2 and (bottom) ResNet-18 trained with GCP or GAP in terms of (left) loss Lipschitzness and (right) gradient predictiveness. Detailed descriptions and discussions can be found in Section~\ref{sec:Smoothing}.}
	\label{fig:landscape}
\end{figure}  
\subsection{Smoothing Effect of GCP}\label{sec:Smoothing}
To understand the mechanism of GCP, we investigate effect of GCP on optimization landscape of deep CNNs. Inspired by~\cite{DBLP:conf/nips/SanturkarTIM18}, we explore effect of GCP on  stability of optimization loss (i.e., Lipschitzness) and stability of gradients (i.e., predictiveness). Specifically, for examining stability of optimization loss, we measure how  loss $\mathcal{L}$ changes along direction of current gradient at each training step. Given the input $\mathbf{X}$, variation of optimization loss is calculated as
\begin{align}\label{variation_loss}
\vartriangle_{l} = \mathcal{L}(\mathbf{X} + \eta_{l}\nabla\mathcal{L}(\mathbf{X})), \eta_{l} \in [a, b],
\end{align}
where $\nabla\mathcal{L}(\mathbf{X})$ indicates the gradient of loss with respect to the input $\mathbf{X}$, and $\eta_{l}$ is step size of gradient descent. Clearly, smaller variation (range) of $\vartriangle_{l}$ indicates that optimization landscape is smoother and more easily controlled during training process~\cite{DBLP:conf/nips/SanturkarTIM18}. As shown in~\cite{DBLP:conf/nips/SanturkarTIM18}, range of $\vartriangle_{l}$ in Eqn.~(\ref{variation_loss}) reflects the Lipschitzness of optimization loss.  

To examine the stability of gradients, we measure how the gradient of loss $\mathcal{L}$ changes by computing the Euclidean distance between the gradient of loss and gradients along the original gradient direction on an interval of step sizes. Thus, the gradient predictiveness can be formulated as
\begin{align}\label{variation_gradient}
\vartriangle_{g} = \|\nabla\mathcal{L}(\mathbf{X})- \nabla\mathcal{L}(\mathbf{X}+\eta_{g}\nabla\mathcal{L}(\mathbf{X}))\|_{2}, \eta_{g} \in [a, b],
\end{align}
where $\eta_{g}$ is step size. Similar to the stability of optimization loss, smaller range of $\vartriangle_{g}$ implies that gradient is more insensitive to step size, having better gradient predictiveness. The loss Lipschitzness and gradient predictiveness greatly affect optimization of deep CNNs, i.e., smoothness of landscape and robustness to hyper-parameters.  

According to the discussions above, we train the networks with GCP or GAP while comparing their loss Lipschitzness and gradient predictiveness to analyze smoothing effect of GCP. Without loss of generality, we employ the widely used MobileNetV2~\cite{DBLP:conf/cvpr/SandlerHZZC18} and ResNet-18~\cite{He_2016_CVPR} as backbone models, and train them on large-scale ImageNet~\cite{imagenet_cvpr09}. For training MobileNetV2 and ResNet-18 with GCP, following~\cite{LiXWZ17}, we reduce dimension of the last convolution activations to 256, and train the networks using stochastic gradient descent (SGD) with the same hyper-parameters in~\cite{DBLP:conf/cvpr/SandlerHZZC18} and~\cite{He_2016_CVPR}. Besides, we do not use $dropout$ operation to avoid randomness for MobileNetV2, and discard down-sampling operation in $conv5\_x$ for ResNet-18. The left endpoints in the ranges of $\vartriangle_{l}$ and $\vartriangle_{g}$ are set to the initial learning rates and right endpoints are as large as possible while ensuring stable training of GAP. As such, $\eta_{l}\,(\eta_{g})\in[0.045,\,1.5]$ and $\eta_{l}\,(\eta_{g})\in[0.1,\,75]$ are for MobileNetV2 and  ResNet-18, respectively.


The behaviors of MobileNetV2 with 500K steps ($\sim$37 epochs) and ResNet-18  with 100K steps ($\sim$20 epochs) trained with GAP and GCP in terms of loss Lipschitzness and gradient predictiveness are shown in top and bottom of Figure~\ref{fig:landscape}, respectively. For both MobileNetV2 and ResNet-18, we observe that the networks with GCP have smaller variations of the optimization loss (i.e., $\vartriangle_{l}$) than GAP-based ones, while optimization losses of the networks with GCP are consistently lower than those trained with GAP. These results demonstrate that GCP can improve Lipschitzness of optimization loss while converging faster under the same setting with GAP-based ones. Meanwhile, for changes of the gradient,  $\vartriangle_{g}$ of the networks with GCP is more stable than GAP-based ones, suggesting the networks with GCP have better gradient predictiveness. In addition, the jumps of both loss landscape and gradient predictiveness for GAP with MobileNetV2 indicate that variations of loss and gradient are considerably large, suggesting varying step sizes are likely to drive the loss uncontrollably higher. In contrast, the variations of one with GCP are fairly small and consistent, suggesting GCP helps for stable training. \emph{In a nutshell, GCP has the ability to smoothen optimization landscape of deep CNNs and improve gradient predictiveness.}

\subsection{Connection to Second-order Optimization}

Furthermore, we analyze back-propagation (BP) of GCP to explore its effect on optimization of deep CNNs. Let $\mathbf{X}$ be output of the last convolution layer, the gradient of loss $\mathcal{L}$ with respect to $\mathbf{X}$ for GAP layer can be computed as $\frac{\partial \mathcal{L}}{\partial \mathbf{X}} = \mathbf{C}^{T}\frac{\partial \mathcal{L}}{\partial \mathbf{Z}_{\text{GAP}}}$, where $\mathbf{Z}_{\text{GAP}} = \sum_{i=1}^{N}\mathbf{X}_{i}$ and $\mathbf{C}$ is a constant matrix. To update weights $\mathbf{W}$ of the last convolution layer, gradient descent is performed as follows:
\begin{align}\label{SGD_GAP}
\mathbf{W}_{t+1} \longleftarrow  \mathbf{W}_{t} - \eta\nabla_{\mathbf{W}}\mathcal{L}_{t},
\end{align}
where $\nabla_{\mathbf{W}}\mathcal{L}_{t}=\frac{\partial \mathcal{L}}{\partial \mathbf{X}}\frac{\partial \mathbf{X}}{\partial \mathbf{W}_{t}}=\mathbf{C}^{T}\frac{\partial \mathcal{L}}{\partial \mathbf{Z}_{\text{GAP}}}\frac{\partial \mathbf{X}}{\partial \mathbf{W}_{t}}$, and $t$ indicates $t$-th iteration.

Let $\mathbf{Z}_{\text{GCP}}=(\mathbf{X}^{T}\mathbf{J}\mathbf{X})^{\frac{1}{2}}$. The derivative of loss with respect to $\mathbf{X}$ for GCP layer can be written as
\begin{align}\label{SGD_GCP}
&\frac{\partial \mathcal{L}}{\partial \mathbf{X}} = 2\mathbf{J}\mathbf{X} \bigg[\mathbf{U}\Big(\Big(\mathbf{K}^{T}\circ \Big(\mathbf{U}^{T}2\Big(\frac{\partial \mathcal{L}}{\partial \mathbf{Z}_{\text{GCP}}}\Big)_{\mathrm{sym}}\mathbf{U}\boldsymbol{\Lambda}^{\frac{1}{2}}\Big)\Big) \nonumber \\ 
&+\Big(\frac{1}{2}\boldsymbol{\Lambda}^{-\frac{1}{2}}\mathbf{U}^{T}\frac{\partial \mathcal{L}}{\partial\mathbf{Z}_{\text{GCP}}}\mathbf{U}\Big)_{\mathrm{diag}}\Big)\mathbf{U}^{T}\bigg]_{\mathrm{sym}},
\end{align}
where  $\mathbf{U}$ and $\boldsymbol{\Lambda}$ are the matrix of eigenvectors and the diagonal matrix of eigenvalues of sample covariance of $\mathbf{X}$, and $\mathbf{K}$ is a mask matrix associated with eigenvalues. Here, $\circ$ denotes matrix Hadamard product; $(\cdot)_{\mathrm{sym}}$ and $(\cdot)_{\mathrm{diag}}$ indicate matrix symmetrization and diagonalization, respectively. More details can refer to~\cite{IonescuVS15,Wang_2017_CVPR,LiXWZ17}. With some assumptions and simplification, Eqn.~(\ref{SGD_GCP}) can be trimmed as  
\begin{align}\label{SGD_GCP3}
\frac{\partial \mathcal{L}}{\partial \mathbf{X}} \approx 2\mathbf{J}\mathbf{X} \Big(2\mathbf{K}^{T}\circ\boldsymbol{\Lambda}^{\frac{1}{2}}+\frac{1}{2}\boldsymbol{\Lambda}^{-\frac{1}{2}}\Big)\frac{\partial\mathcal{L}}{\partial \mathbf{Z}_{\text{GCP}}}.
\end{align}
Details of Eqn.~(\ref{SGD_GCP}) and Eqn.~(\ref{SGD_GCP3}) can be found in the supplemental file. By substituting Eqn.~(\ref{SGD_GCP3}) into Eqn.~(\ref{SGD_GAP}), we can approximatively update the weights of convolution as
\begin{align}\label{SGD_GCP4}
\mathbf{W}_{t+1} \longleftarrow  \mathbf{W}_{t} - \mathbf{F}^{-1}\frac{\partial \mathcal{L}}{\partial \mathbf{Z}_{\text{GCP}}}\frac{\partial \mathbf{X}}{\partial \mathbf{W}_{t}},
\end{align}
where $\mathbf{F}^{-1}=\eta2\mathbf{J}\mathbf{X} \Big(2\mathbf{K}^{T}\circ\boldsymbol{\Lambda}^{\frac{1}{2}}+\frac{1}{2}\boldsymbol{\Lambda}^{-\frac{1}{2}}\Big)$. 

\begin{table}
	\begin{center}
		\footnotesize
		\renewcommand\arraystretch{1.72}
		\begin{tabular}{|l|c|c|}
			\hline
			Method & Gradient & Remark \\
			\hline\hline
			GAP	& $\eta\mathbf{C}^{T}\frac{\partial \mathcal{L}}{\partial \mathbf{Z}_{\text{GAP}}}\frac{\partial \mathbf{X}}{\partial \mathbf{W}_{t}}$ & $\mathbf{C}$ is a constant matrix \\
			\hline
			GCP	& $\approx\mathbf{F}^{-1}\frac{\partial \mathcal{L}}{\partial \mathbf{Z}_{\text{GCP}}}\frac{\partial \mathbf{X}}{\partial \mathbf{W}_{t}}$ & \tabincell{c}{$\mathbf{F}^{-1}=\eta2\mathbf{J}\mathbf{X} \Big(2\mathbf{K}^{T}\circ\boldsymbol{\Lambda}^{\frac{1}{2}} $ \\ $+  \frac{1}{2}\boldsymbol{\Lambda}^{-\frac{1}{2}}\Big)$}  \\
			\hline
			K-FAC~\cite{DBLP:conf/cvpr/OsawaTUNYM19}& $\mathbf{H}^{-1}\mathbf{C}^{T}\frac{\partial \mathcal{L}}{\partial \mathbf{Z}_{\text{GAP}}}\frac{\partial \mathbf{X}}{\partial \mathbf{W}_{t}}$ & $\mathbf{H}^{-1}=\eta \big(\frac{\partial \mathcal{L}}{\partial \mathbf{X}}\big)^{-1}\otimes\widehat{\mathbf{X}}^{-1}$\\
			\hline
		\end{tabular}
	\end{center}
	\caption{Comparison of gradients involved in GAP, GCP and GAP with K-FAC~\cite{DBLP:conf/cvpr/OsawaTUNYM19}. $\otimes$ indicates Kronecker product.}
	\label{table:BP}
\end{table}

Previous works~\cite{DBLP:conf/icml/Martens10,DBLP:conf/icml/MartensG15,DBLP:conf/cvpr/OsawaTUNYM19} show second-order optimization (i.e., $\mathbf{W}_{t+1} \longleftarrow  \mathbf{W}_{t} - \mathbf{H}^{-1}\nabla\mathcal{L}_{t}$) can speed up training of deep neural networks. However, computation of inverse of Hessian matrix ($\mathbf{H}^{-1}$) is usually very expensive and is sensitive to noise. Therefore, many methods~\cite{DBLP:conf/icml/Martens10,DBLP:conf/icml/MartensG15,DBLP:journals/jmlr/AgarwalBH17} are proposed to approximate $\mathbf{H}^{-1}$. Recently, K-FAC~\cite{DBLP:conf/icml/MartensG15,DBLP:conf/cvpr/OsawaTUNYM19} based on an accurate approximation of the Fisher information matrix has proven to be effective in optimizing deep CNNs, which approximates $\mathbf{H}^{-1}$ using Kronecker product between inverse of input of convolution layer (i.e., $\widehat{\mathbf{X}}^{-1}$) and inverse of gradient of the loss with respect to the output (i.e., $\big(\frac{\partial \mathcal{L}}{\partial \mathbf{X}}\big)^{-1}$). The gradients involved in GAP, GCP and GAP with K-FAC are compared in Table~\ref{table:BP}, where the trimmed BP of GCP~(\ref{SGD_GCP3}) shares some similar philosophy with K-FAC. The key difference is that $\mathbf{F}^{-1}$ is computed by the output $\mathbf{X}$ and its eigenvalues, while $\mathbf{H}^{-1}$ is approximated by the input $\widehat{\mathbf{X}}$ and the gradient $\frac{\partial \mathcal{L}}{\partial \mathbf{X}}$. The experiments in Section~\ref{sec:fastconv} show that ResNet-50 with GCP uses less training epochs to achieve result matching that of K-FAC~\cite{DBLP:conf/cvpr/OsawaTUNYM19}, which may indicate that BP of GCP is a potential alternative of pre-conditioner for Hessian matrix. 

\section{Merits Benefited from GCP}

\begin{figure*}
	\centering
	\subfigure[MobileNetV2]{
		\centering
		\includegraphics[width=0.3\textwidth]{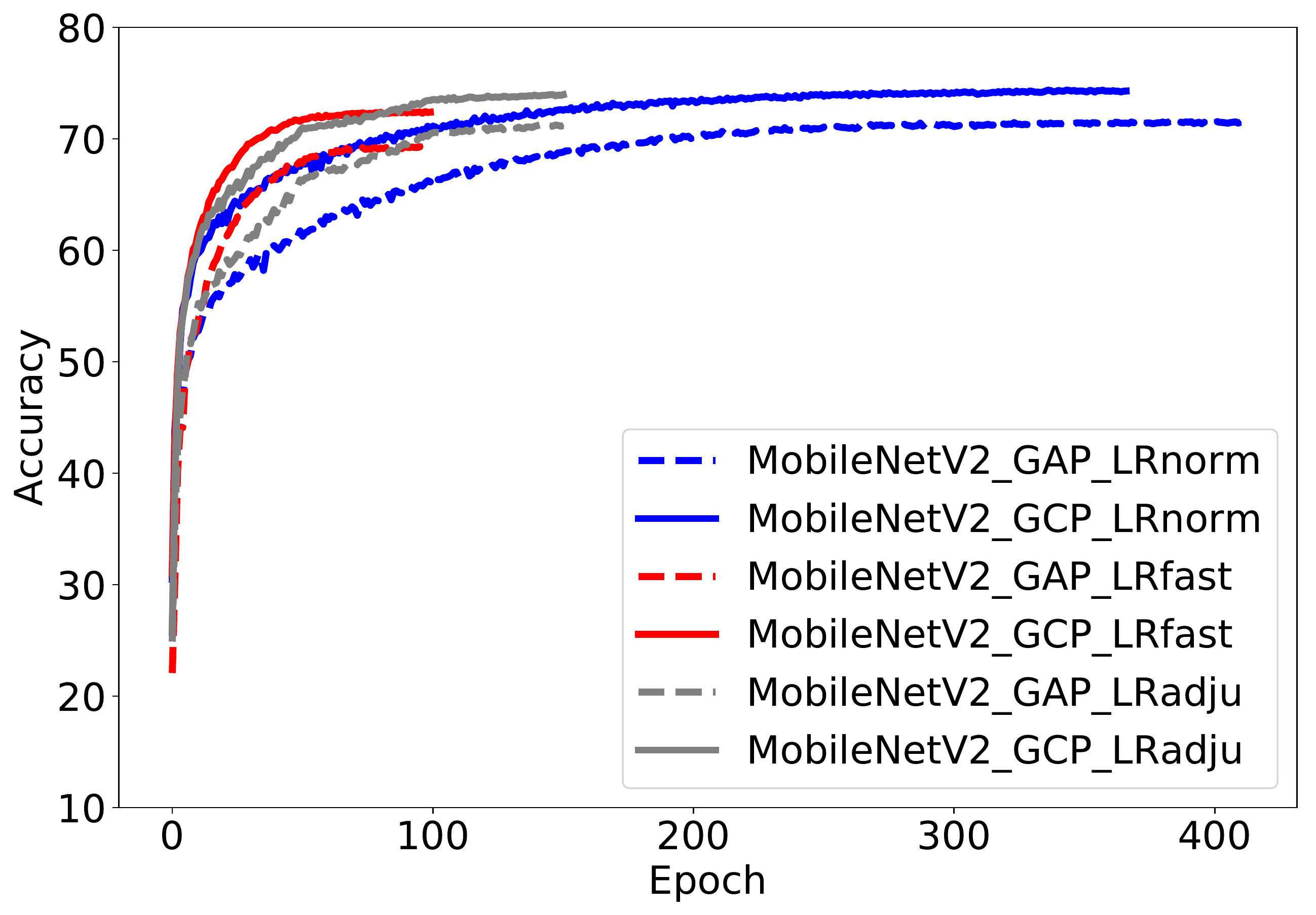} 
		\label{fig:MobileNetV2} 
	}
	\subfigure[ShuffleNet V2]{
		\centering
		\includegraphics[width=0.3\textwidth]{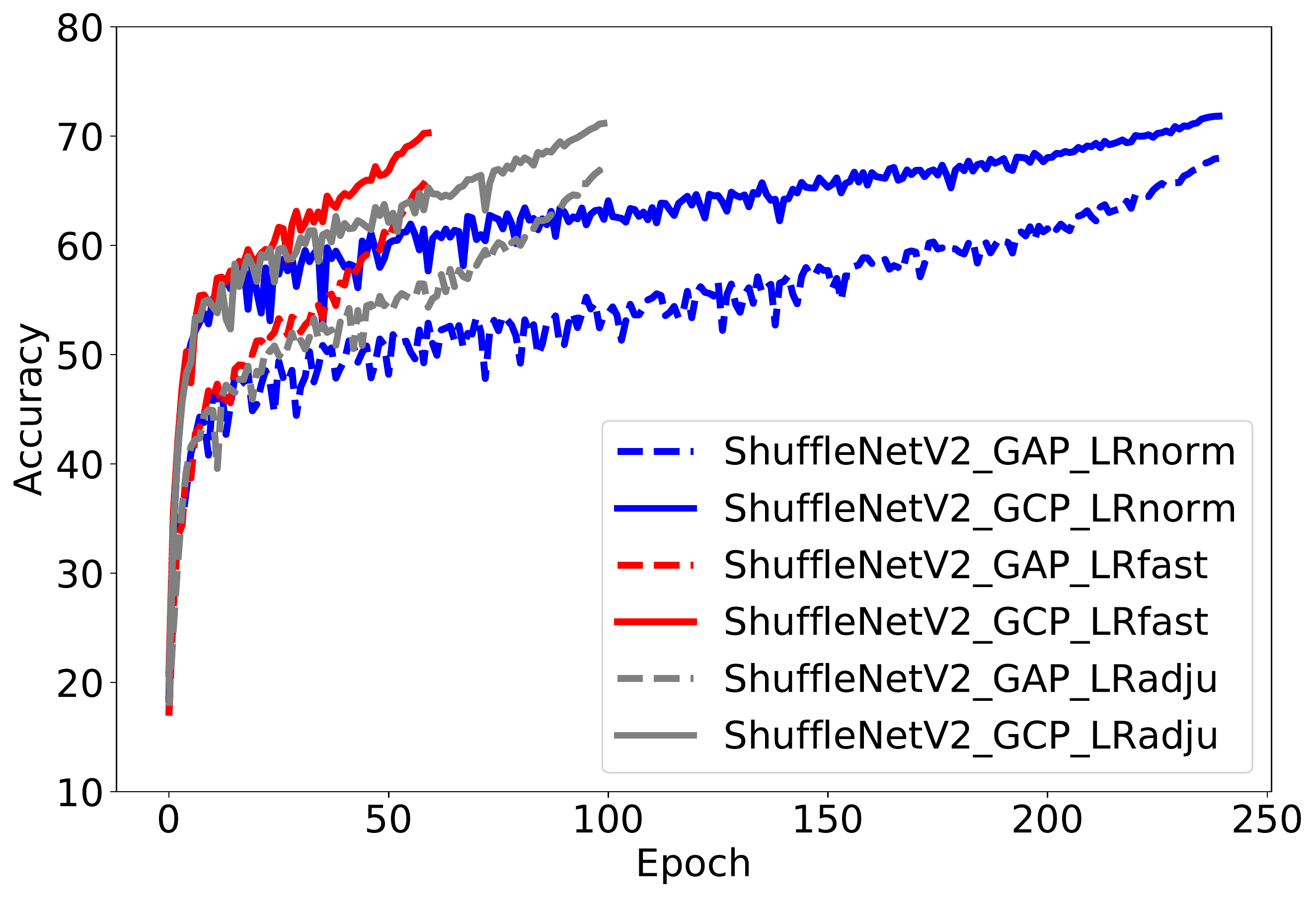}
		\label{fig:ShuffleNetV2} 
	}
	\subfigure[ResNet-18]{
		\centering
		\includegraphics[width=0.3\textwidth]{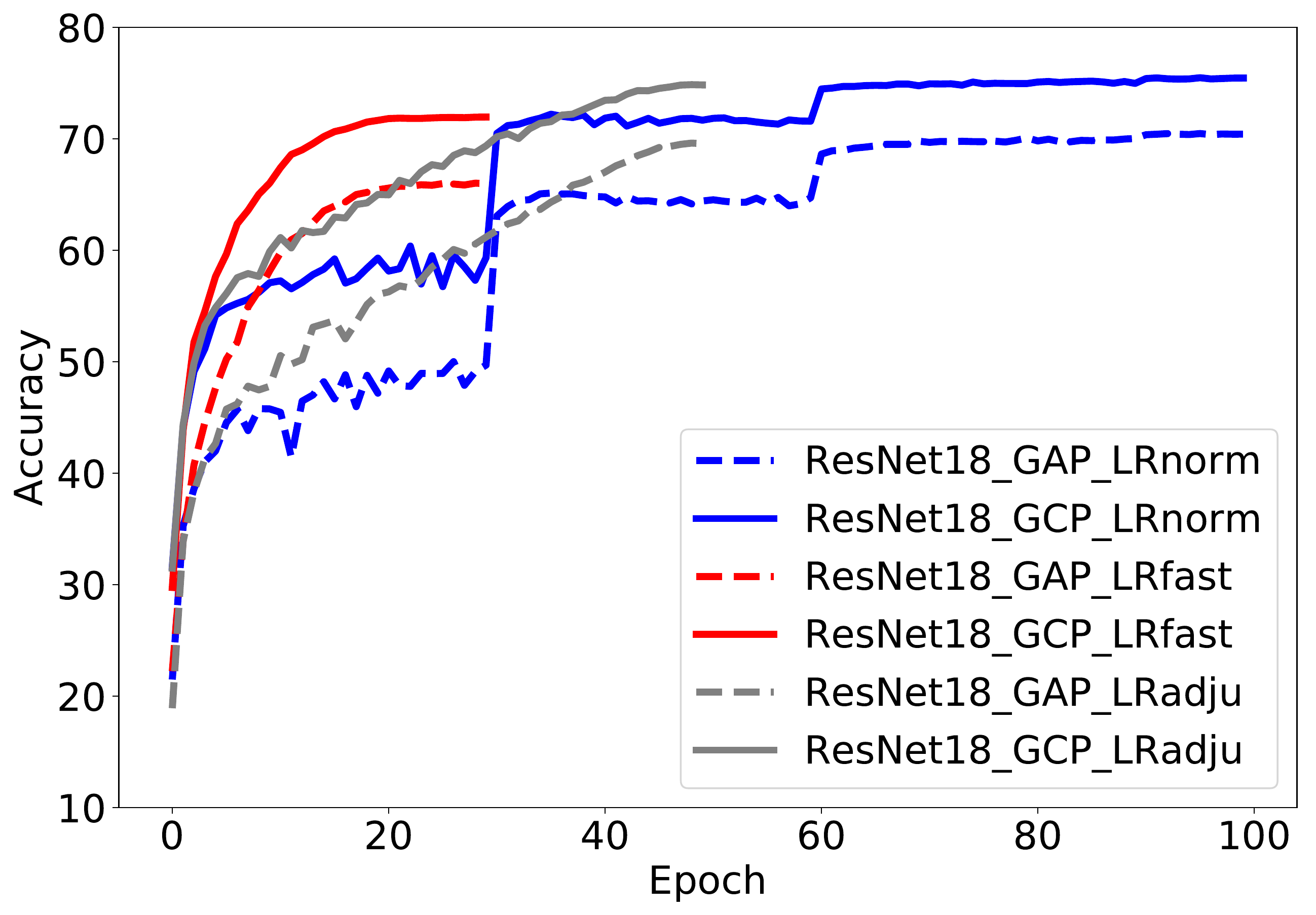} 
		\label{fig:resnet18}
	}
	
	\subfigure[ResNet-34]{
		\centering
		\includegraphics[width=0.3\textwidth]{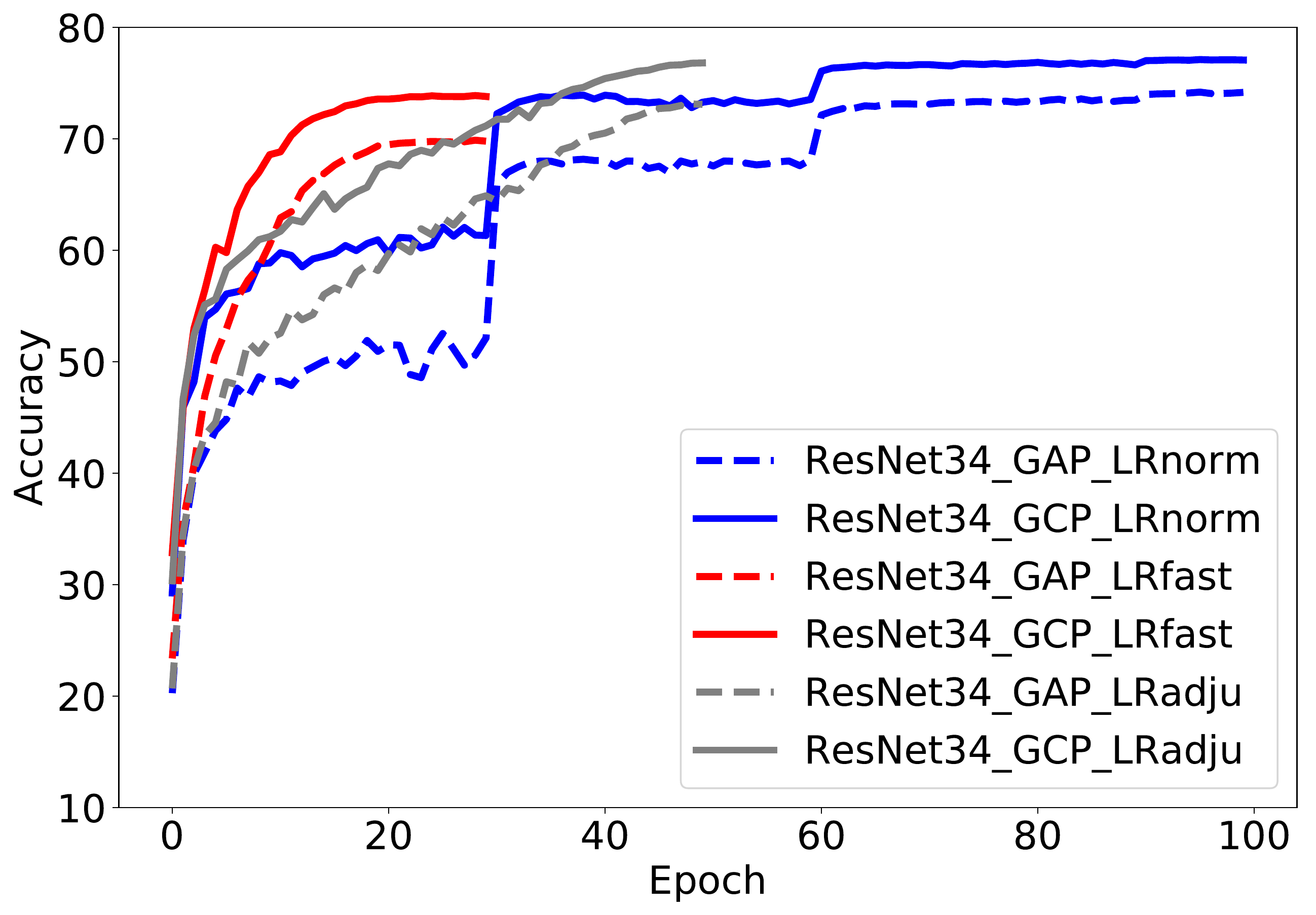} 
		\label{fig:resnet34} 
	}    
	\subfigure[ResNet-50]{
		\centering
		\includegraphics[width=0.3\textwidth]{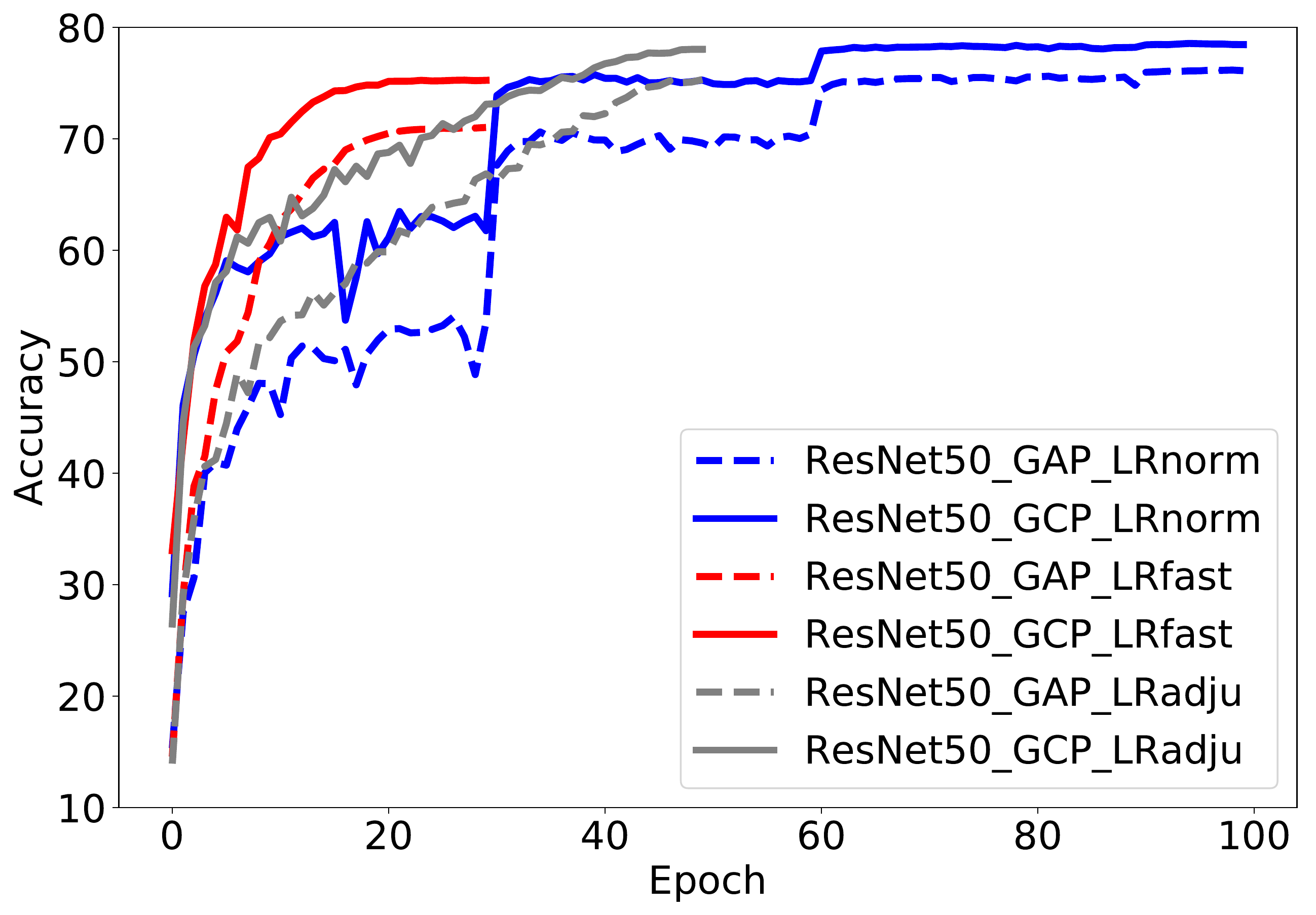}
		\label{fig:resnet50} 
	}
	\subfigure[ResNet-101]{
		\centering
		\includegraphics[width=0.3\textwidth]{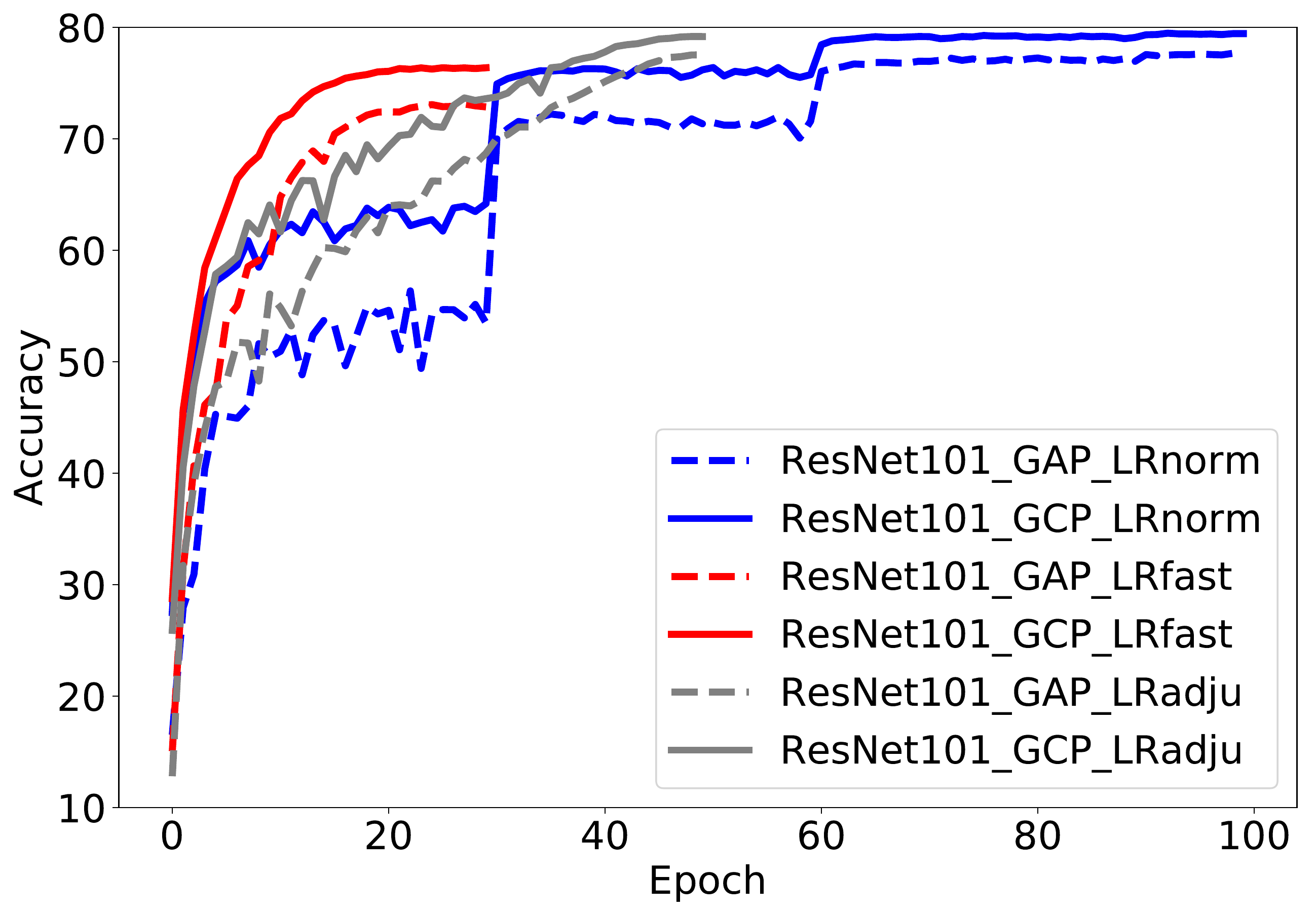} 
		\label{fig:resnet101}
	}
	\caption{Convergence curves of different deep CNN models trained with GAP and GCP under various settings of learning rates (i.e., $LR_{norm}$, $LR_{fast}$ and $LR_{adju}$) on ImageNet.}
	\label{fig:fastconv}
\end{figure*}

\begin{figure}[t]
	\begin{center}
		\includegraphics[width=1.0\linewidth]{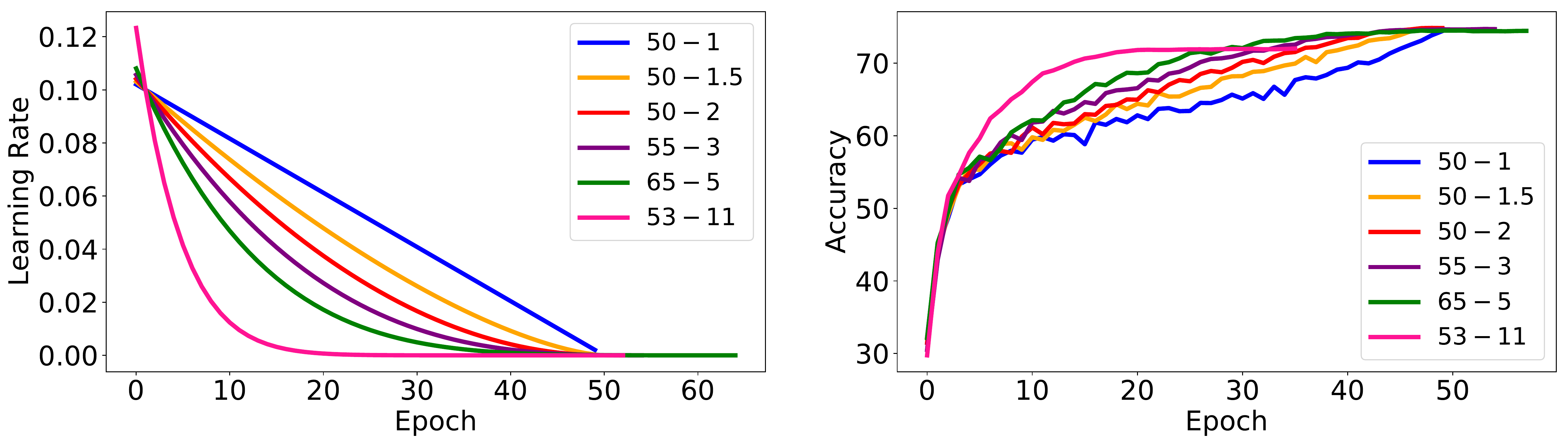}
	\end{center}
	\caption{(Left) Curves of various settings ($\rho-e_f$) of $lr$ and (Right) their corresponding convergence curves using ResNet-18 as backbone model on ImageNet. Here $e_f$ and  $\rho$ denote the final epoch and power of the polynomial decay in Eqn.~(\ref{equ:lr_poly}).}
	\label{fig:lr}
\end{figure} 

In previous section, we explore the effect of GCP on deep CNNs from an optimization perspective. Specifically, we show smoothing effect of GCP on  optimization landscape, and discuss the connection between GCP and second-order optimization. Furthermore,  these findings can also account for several merits delivered by GCP for training deep CNNs that have not been recognized previously or fully explored, including significant acceleration of network convergence, stronger robustness to distorted examples generated by image corruptions and perturbations, and good generalization ability to different vision tasks. In this work, all programs are implemented by Pytorch package\footnote{\url{https://github.com/ZhangLi-CS/GCP_Optimization}}, and run on a workstation with four NVIDIA RTX 2080 Ti GPUs. The detailed descriptions are given as follows. 

\begin{table*}
	\begin{center}
		\footnotesize	
		\begin{tabular}{|l|c|c|c|c|l|l|l|}
			\hline
			Backbone Model & Pooling Method & $lr$ & BS & Training  Epochs & Matching Epoch & Top-1 Accuracy & Top-5 Accuracy \\
			\hline\hline
			\multirow{2}{*}{MobileNetV2} & GAP & $LR_{norm}$ & 96 & 400 & N/A & 71.58 & 90.30 \\
			& GCP& $LR_{adju}$ & 192 & 150 & 68$_{(\downarrow332)}$ & 73.97 $_{(\uparrow2.39)}$ & 91.54$_{(\uparrow1.24)}$  \\
			\hline
			\multirow{2}{*}{ShuffleNet V2} & GAP & $LR_{norm}$ & 1,024 & 240 & N/A & 67.96  & 87.84 \\
			& GCP& $LR_{adju}$& 1,024& 100 & 78$_{(\downarrow162)}$  & 71.17$_{(\uparrow3.21)}$ & 89.74$_{(\uparrow1.90)}$ \\
			\hline
			\multirow{2}{*}{ResNet-18} & GAP & $LR_{norm}$ & 256 & 100 & N/A & 70.47 & 89.62 \\
			& GCP& $LR_{adju}$ & 256 & 50 & 32$_{(\downarrow68)}$  & 74.86$_{(\uparrow4.39)}$  & 91.81$_{(\uparrow2.19)}$\\
			\hline
			\multirow{2}{*}{ResNet-34} & GAP & $LR_{norm}$ & 256 & 100 & N/A & 74.19 & 91.61 \\
			& GCP& $LR_{adju}$ & 256 & 50 & 38$_{(\downarrow62)}$ & 76.81$_{(\uparrow2.62)}$ & 93.09$_{(\uparrow1.48)}$ \\
			\hline
			\multirow{4}{*}{ResNet-50} & GAP & $LR_{norm}$ & 256 & 100 & N/A & 76.17 & 92.93 \\
			& GCP& $LR_{adju}$ & 256& 50 & 40$_{(\downarrow60)}$ & 78.03$_{(\uparrow1.86)}$ & 93.95$_{(\uparrow1.02)}$ \\
			\cline{2-8}
			& GAP& K-FAC~\cite{DBLP:conf/cvpr/OsawaTUNYM19}$^{\star}$ & 4,096 & 35 & N/A & 75.10$_{(\downarrow1.07)}$ & N/A \\
			& GCP& $LR_{fast}$ & 256& 30 & 20$_{(\downarrow15)}$ & 75.31$_{(\downarrow0.86)}$ & 92.15$_{(\downarrow0.78)}$ \\
			\hline
			\multirow{2}{*}{ResNet-101} & GAP & $LR_{norm}$ & 256 & 100 & N/A & 77.67  & 93.89 \\
			& GCP& $LR_{adju}$ & 256 & 50 & 41$_{(\downarrow59)}$ & 79.18$_{(\uparrow1.51)}$ & 94.51$_{(\uparrow0.62)}$ \\
			\hline
		\end{tabular}
	\end{center}
	\caption{Comparison of different CNNs trained with GAP using $LR_{norm}$ and those trained with GCP using $LR_{adju}$ on ImageNet. `Matching Epoch' indicates that at which the networks with GCP achieve comparable performance with the corresponding networks with GAP undergoing full training epochs. $^{\star}$: The result of K-FAC is duplicated from the original paper~\cite{DBLP:conf/cvpr/OsawaTUNYM19}.}
	\label{table:fastconv}
\end{table*}

\begin{table*}[!t]
	\begin{minipage}[!t]{0.75\columnwidth}
		\begin{center}
			\footnotesize
			\begin{tabular}{|l|c|c|}
				\hline
				\multicolumn{2}{|c|}{Method}            &   \multirow{2}*{Setting of $lr$ }                     \\
				\cline{1-2}
				Backbone                      & $lr$     & ~                     ~                                            \\
				\hline
				\hline
				\multirow[b]{3}*{MobileNet V2}    & $LR_{norm}$                      & $0.045 \times 0.98^{e}$         \\   
				\cline{2-3}
				~                             & $LR_{fast}$                      & $0.06 \times 0.92^{e}$               \\
				\cline{2-3}                                                                                                  
				~                             & $LR_{adju}$                  & 
				$
				\begin{cases}   
				linear(6e^{-2},1e^{-3},0), \\ 
				linear(1e^{-2},1e^{-4},50), \\  
				linear(1e^{-3},1e^{-5},100) 
				\end{cases} 
				$                                
				\\
				\hline                                                                                                       
				\multirow{3}*{ShuffleNetV2}   & $LR_{norm}$                    & \multirow{3}*{$0.5 \times (1-\frac{step}{t\_step})$}       \\
				\cline{2-2}                                                                                                  
				~                             & $LR_{fast}$                     & ~                                       \\
				\cline{2-2}                                                                                                  
				~                             & $LR_{adju}$                    & ~                                        \\
				\hline                                                                                                       
				\multirow{3}*{ResNet}         & $LR_{norm}$                     & $0.1^{(e//30) + 1}$                 \\
				\cline{2-3}                                                                                                  
				~                             & $LR_{fast}$                      & $0.1 \times (1-\frac{e-1}{52}) ^ {11}$   \\
				\cline{2-3}                                                                                                  
				~                             & $LR_{adju}$                     & $0.1 \times (1-\frac{e-1}{49}) ^ {2}$    \\
				\hline 
			\end{tabular}
		\end{center}
		\caption{Detailed settings of $lr$ for various CNNs.}
		\label{table:lr}
	\end{minipage}
	\begin{minipage}[!t]{1.54\columnwidth}
		\begin{center}
			\footnotesize
			\begin{tabular}{|l|l|l|l|l|}
				\hline
				\multirow{2}{*}{Method} & \multicolumn{2}{c|}{IMAGENET-C} & \multicolumn{2}{c|}{IMAGENET-P} \\
				\cline{2-5}
				&  mCE              & Relative         mCE                 & mFP              &mT5D               \\
				\hline                                                                      \hline                
				MobileNetV2 + GAP   & 87.1             & 114.9             & 79.8             & 96.5                \\
				MobileNetV2 + GCP   & 81.7$_{(\downarrow5.4)}$   & 110.6$_{(\downarrow4.3)}$   & 64.3$_{(\downarrow15.5)}$  & 87.6$_{(\downarrow8.9)}$      \\
				\hline                                                                                            
				ShuffleNet V2 + GAP & 92.7             & 126.7             & 94.7             & 108.2               \\
				ShuffleNet V2 + GCP & 85.2$_{(\downarrow7.5)}$   & 112.6$_{(\downarrow14.1)}$  & 75.2$_{(\downarrow19.5)}$  & 95.5$_{(\downarrow12.7)}$     \\
				\hline                                                                                            
				ResNet-18 + GAP     & 84.7             & 103.9             & 72.8             & 87.0               \\
				ResNet-18 + GCP     & 76.3$_{(\downarrow8.4)}$   & 101.3$_{(\downarrow2.6)}$   & 53.2$_{(\downarrow19.6)}$  & 77.1$_{(\downarrow9.9)}$   \\
				\hline                                                                                            
				ResNet-34 + GAP     & 77.9             & 98.7              & 61.7             & 79.5               \\
				ResNet-34 + GCP     & 72.4$_{(\downarrow5.5)}$   & 96.9$_{(\downarrow1.8)}$    & 47.7$_{(\downarrow14.0)}$  & 72.4$_{(\downarrow7.1)}$     \\
				\hline                                                                                            
				ResNet-50 + GAP     & 76.7             & 105.0             & 58.0             & 78.3               \\
				ResNet-50 + GCP     & 70.7$_{(\downarrow6.0)}$   & 97.9$_{(\downarrow7.1)}$    & 47.5$_{(\downarrow10.5)}$  & 74.6$_{(\downarrow3.7)}$     \\
				\hline                                                                                            
				ResNet-101 + GAP    & 70.3             & 93.7              & 52.6             & 73.9               \\
				ResNet-101 + GCP    & 65.5$_{(\downarrow4.8)}$   & 89.1$_{(\downarrow4.6)}$    & 42.1$_{(\downarrow10.5)}$  & 68.3$_{(\downarrow5.6)}$      \\
				\hline                                                                
			\end{tabular}
		\end{center}
		\caption{Comparison of GAP and GCP on IMAGENET-C and IMAGENET-P.}
		\label{adversarial}
	\end{minipage}
\end{table*}

\subsection{Acceleration of Network Convergence}\label{sec:fastconv}
It is well known that training of deep CNNs is a time-consuming process, requiring  a mass of computing resources. Therefore, acceleration of network convergence plays a key role in fast training of deep CNNs, especially for large-scale training datasets. Previous study~\cite{LiXWG18} empirically shows the networks trained with GCP converge faster than GAP-based ones, which can be explained by our findings. Specifically, Lipschitzness improvement~\cite{DBLP:conf/nips/SanturkarTIM18,DBLP:conf/iclr/MiyatoKKY18} and connection to second-order optimization~\cite{DBLP:conf/cvpr/OsawaTUNYM19} brought by GCP can accelerate convergence of networks. Here, we further show deep CNNs with GCP can support rapid decay of learning rates for significantly improving convergence speed, due to its ability of smoothing optimization landscape. To verify it, we conduct extensive experiments using six representative deep CNN architectures on ImageNet, including MobileNetV2~\cite{DBLP:conf/cvpr/SandlerHZZC18}, ShuffleNet V2~\cite{DBLP:conf/eccv/MaZZS18} and ResNets~\cite{He_2016_CVPR} of 18, 34, 50 and 101 layers.

Specifically, we train all the networks with GAP and GCP by varying the settings of learning rates ($lr$). Firstly, we use the setting of $lr$ in each original paper, which is indicted by $LR_{norm}$. As shown in~\cite{DBLP:conf/cvpr/OsawaTUNYM19}, ResNet-50 converges much faster using second-order optimization (K-FAC), in which $lr$ is scheduled by polynomial decay, i.e.,
\begin{align}\label{equ:lr_poly}
\ell_{e} = \ell_{0} \times \bigg(1-\frac{e-e_s}{e_f-e_s}\bigg)^{\rho},
\end{align}
where $\ell_{0}$ is the initial $lr$; $e$, $e_s$ and $e_f$ indicate $e$-th, the start and the final epochs, respectively. The parameter $\rho$ controls decay rate. Inspired by~\cite{DBLP:conf/cvpr/OsawaTUNYM19}, we employ the setting of $lr$ in Eqn.~(\ref{equ:lr_poly}) for fast training of ResNets. To determine parameter $\rho$, we set $\ell_{0}=0.1$ and train ResNet-18 with GCP on ImageNet using various $\rho$ while controlling $e_f$ be less than 65, which is consistent with~\cite{DBLP:conf/cvpr/OsawaTUNYM19}. Figure~\ref{fig:lr} illustrates the curve of each $lr$ and its corresponding convergence. Clearly, larger $\rho$ leads to faster convergence but lower accuracy. The setting with $(\rho=11, e_f=53)$ has the fastest convergence, denoted by $LR_{fast}$ hereafter, which converges within 30 epochs. Among them, the setting with $(\rho=2, e_f=50)$ makes the best trade-off between convergence speed and classification accuracy, denoted by $LR_{adju}$.

For MobileNetV2, the original $lr$ is scheduled by exponential decay (i.e., $0.045 \times 0.98^{e}$), and we set $LR_{fast}$ by decreasing base number while increasing the initial $lr$, i.e., $0.06 \times 0.92^{e}$. $LR_{adju}$ is scheduled by a stage-wise linear decay, i.e., $linear(l_s,l_e,n) = l_s-\frac{l_s-l_e}{50}(e-n)$ for $e$-th epoch where $l_s$, $l_e$ and $n$ are initial $lr$, final $lr$ and start epoch in each stage. The $lr$ of ShuffleNet V2 in the original paper is scheduled by a step-wise linear decay, and number of total steps (i.e., $t\_step$) is 3e5 ($\sim$ 240 epochs). For ShuffleNet V2, $LR_{fast}$ and $LR_{adju}$ are set by reducing training epochs to 60 and 100, respectively. The detailed settings of $lr$ are summarized in Table~\ref{table:lr}.

\begin{table}
	\begin{center}
		\footnotesize
		\begin{tabular}{|l|c|c|l|l|}
			\hline
			Backbone & $lr$ & $D$ & Top-1 Acc. & Top-5 Acc. \\
			\hline\hline
			\multirow{4}{*}{MobiNetV2} & \multirow{2}{*}{$LR_{norm}$} & $256$ & 74.36 & 91.90 \\
			&  & $128$ & 73.28 ($\downarrow1.08$) & 91.30 ($\downarrow0.60$)  \\
			\cline{2-5}
			& \multirow{2}{*}{$LR_{adju}$} & $256$ & 73.97 & 91.54 \\
			&  & $128$ & 72.58 ($\downarrow1.39$) & 90.90 ($\downarrow0.64$) \\
			\hline
			\multirow{4}{*}{ResNet-50} & \multirow{2}{*}{$LR_{norm}$} & $256$ & 78.56 & 94.16 \\
			&  & $128$ & 78.21 ($\downarrow 0.35$) & 93.93 ($\downarrow 0.23$) \\
			\cline{2-5}
			& \multirow{2}{*}{$LR_{adju}$} & $256$ & 78.03 & 93.95 \\
			&  & $128$ & 77.64 ($\downarrow 0.39$) &  93.67 ($\downarrow 0.28$)\\
			\hline
		\end{tabular}
	\end{center}
	\caption{Results of MobileNetV2 and ResNet-50 trained with GCP using different $lr$ and dimensions ($D$) of input on ImageNet.}
	\label{table:lim_HD}
\end{table}

The convergence curves of different networks trained with GAP and GCP under various settings of $lr$ are illustrated in Figure~\ref{fig:fastconv}, from which we have the following observations. (1) Comparing with the networks trained under $LR_{norm}$, those trained under the settings of $LR_{fast}$ and $LR_{adju}$ suffer from performance degradation. However, performance degradations of the networks with GCP are less than those based on GAP, especially under $LR_{fast}$. (2) The networks trained with GCP using $LR_{fast}$ achieve better or matching results using only about 1/4 of the number of epochs for training networks with GAP under $LR_{norm}$. 

Moreover, we compare the networks trained with GAP under $LR_{norm}$ (i.e., the original settings) and those trained with GCP using $LR_{adju}$. According to the results in Table~\ref{table:fastconv}, we make a summary as follows. (1) Comparing to the networks trained with GAP under $LR_{norm}$, those trained with GCP under $LR_{adju}$ achieve higher accuracies while using less training epochs. (2) The networks with GCP obtain matching or comparable accuracies to GAP-based ones using much less training epochs, especially for lightweight CNN models, i.e., MobileNetV2 and ShuffleNet V2. For example, MobileNetV2 with GCP achieves matching accuracies to GAP-based one using only 68 epochs, while the latter one needs about 400 epochs. (3) Comparing to the second-order optimization method K-FAC~\cite{DBLP:conf/cvpr/OsawaTUNYM19}, GCP obtains moderate accuracy gain using less training epochs, while achieving comparable accuracies with K-FAC using only 20 epochs. Furthermore, GCP is easier to implement. \emph{The extensive experiments above strongly support our finding: GCP can significantly speed up convergence of deep CNNs with rapid decay of learning rates.}

Additionally, we assess the effect of dimension of covariance representations (COV-Reps) on behavior of convergence using MobileNetV2 (MobiNetV2) and ResNet-50 on ImageNet. If the dimension of input features is $D$, GCP will output a $D(D+1)/2$-dimensional COV-Reps. Here, we set $D$ to 256 (the default setting) and 128, and train the networks under the settings of $LR_{norm}$ and $LR_{adju}$, respectively. The results are given in Table~\ref{table:lim_HD}, from which one can see that lower-dimensional COV-Reps still allow faster convergence of deep CNNs, but suffer from larger performance degradation in the case of faster convergence (i.e., $LR_{adju}$). This indicates that dimension of COV-Reps has a nontrivial effect on the behavior of convergence. Therefore, how to compress COV-Reps while preserving merits of high-dimensional ones is an important issue. Albeit many works are proposed to compress COV-Reps~\cite{Gao_2016_CVPR,Kong_Charless_2017_CVPR,DBLP:conf/eccv/YuS18}, they still have not been verified in large-scale scenarios. A potential solution is to learn compact COV-Reps from high-dimensional ones based on knowledge distillation~\cite{DBLP:journals/corr/HintonVD15}, which will be studied in future.

\begin{table*}[t]
	\begin{center}
		\footnotesize
		\begin{tabular}{|l|l|c|l|l|l|l|l|l|}
			\hline
			\multicolumn{1}{|l|}{Backbone Model} & \multicolumn{1}{l|}{Method} & \multicolumn{1}{c|}{Detectors} & \multicolumn{1}{c|}{$AP$} & \multicolumn{1}{c|}{$AP_{50}$} & \multicolumn{1}{c|}{$AP_{75}$} & \multicolumn{1}{c|}{$AP_S$} & \multicolumn{1}{c|}{$AP_M$} & \multicolumn{1}{c|}{$AP_L$} \\
			\hline 
			\hline 
			\multirow{3}*{ResNet-50}  & GAP               & \multirow{6}*{Faster R-CNN}   & 36.4           & 58.2           & 39.2           & 21.8           & 40.0           & 46.2           \\
			\cline{2-2}\cline{4-9}
			~                         & GCP$_{D}$     & ~                             & 36.6$_{(\uparrow0.2)}$ & 58.4$_{(\uparrow0.2)}$ & 39.5$_{(\uparrow0.3)}$ & 21.3$_{(\downarrow0.5)}$& 40.8$_{(\uparrow0.8)}$ & 47.0$_{(\uparrow0.8)}$ \\
			~                         & GCP$_{M}$  & ~                             & \textbf{37.1$_{\mathbf{(\uparrow0.7)}}$} & \textbf{59.1$_{\mathbf{(\uparrow0.9)}}$} & \textbf{39.9$_{\mathbf{(\uparrow0.7)}}$} & \textbf{22.0$_{\mathbf{(\uparrow0.2)}}$} & \textbf{40.9$_{\mathbf{(\uparrow0.9)}}$} & \textbf{47.6$_{\mathbf{(\uparrow1.4)}}$} \\
			\cline{1-2}\cline{4-9}
			\multirow{3}*{ResNet-101} & GAP               & ~                             & 38.7           & 60.6           & 41.9           & 22.7           & 43.2           & 50.4           \\
			\cline{2-2}\cline{4-9}
			~                         & GCP$_{D}$     & ~                             & 39.5$_{(\uparrow0.8)}$ & 60.7$_{(\uparrow0.1)}$ & 43.1$_{(\uparrow1.2)}$ & 22.9$_{(\uparrow0.2)}$ & \textbf{44.1$_{\mathbf{(\uparrow0.9)}}$} & \textbf{51.4$_{\mathbf{(\uparrow1.0)}}$} \\
			~                         & GCP$_{M}$  & ~                             & \textbf{39.6$_{\mathbf{(\uparrow0.9)}}$} & \textbf{61.2$_{\mathbf{(\uparrow0.6)}}$} & \textbf{43.1$_{\mathbf{(\uparrow1.2)}}$} & \textbf{23.3$_{\mathbf{(\uparrow0.6)}}$} & 43.9$_{(\uparrow0.7)}$ & 51.3$_{(\uparrow0.9)}$ \\
			\hline
			\multirow{3}*{ResNet-50}  & GAP               & \multirow{6}*{Mask R-CNN}     & 37.2           & 58.9           & 40.3           & 22.2           & 40.7           & 48.0           \\
			\cline{2-2}\cline{4-9}                                                                                      
			~                         & GCP$_{D}$     & ~                             & 37.3$_{(\uparrow0.1)}$ & 58.8$_{(\downarrow0.1)}$& 40.4$_{(\uparrow0.1)}$ & 22.0$_{(\downarrow0.2)}$& 41.1$_{(\uparrow0.4)}$ & 48.2$_{(\uparrow0.2)}$ \\
			~                         & GCP$_{M}$  & ~                             & \textbf{37.9$_{\mathbf{(\uparrow0.7)}}$} & \textbf{59.4$_{\mathbf{(\uparrow0.5)}}$} & \textbf{41.3$_{\mathbf{(\uparrow1.0)}}$} & \textbf{22.4$_{\mathbf{(\uparrow0.2)}}$} & \textbf{41.5$_{\mathbf{(\uparrow0.8)}}$} & \textbf{49.0$_{\mathbf{(\uparrow1.0)}}$} \\
			\cline{1-2}\cline{4-9}
			\multirow{3}*{ResNet-101} & GAP               & ~                             & 39.4           & 60.9           & 43.3           & 23.0           & 43.7           & 51.4           \\
			\cline{2-2}\cline{4-9}
			~                         & GCP$_{D}$     & ~                             & 40.3$_{(\uparrow0.9)}$ & 61.5$_{(\uparrow0.6)}$ & 44.0$_{(\uparrow0.7)}$ & \textbf{24.1$_{\mathbf{(\uparrow1.1)}}$} & 44.7$_{(\uparrow1.0)}$ & 52.5$_{(\uparrow1.1)}$ \\
			~                         & GCP$_{M}$  & ~                             & \textbf{40.7$_{\mathbf{(\uparrow1.3)}}$} & \textbf{62.0$_{\mathbf{(\uparrow1.1)}}$} & \textbf{44.6$_{\mathbf{(\uparrow1.3)}}$} & 23.9$_{(\uparrow0.9)}$ & \textbf{45.2$_{\mathbf{(\uparrow1.5)}}$} & \textbf{52.9$_{\mathbf{(\uparrow1.5)}}$} \\
			\hline
		\end{tabular}
	\end{center}
	\caption{Object detection results of various deep CNN models using Faster R-CNN and Mask R-CNN on COCO val2017.}
	\label{detection}
\end{table*}

\begin{table}[t]
	\begin{center}
		\footnotesize
		\begin{tabular}{|l|c|c|c|c|c|c|}
			\hline
			Method       & $AP$ & $AP_{50}$ & $AP_{75}$ & $AP_S$ & $AP_M$ & $AP_L$ \\
			\hline
			\hline
			R-50 + GAP                & 34.1 & 55.5 & 36.2 & 16.1 & 36.7 & 50.0 \\
			R-50 + GCP$_{D}$      & 34.2 & 55.3 & 36.4 & 15.8 & 37.1 & 50.1\\
			R-50 + GCP$_{M}$    & \textbf{34.7} &\textbf{56.3} & \textbf{36.8} & \textbf{16.4} & \textbf{37.5} & \textbf{50.6} \\
			\hline
			R-101 + GAP                & 35.9 & 57.7 & 38.4 & 16.8 & 39.1 & 53.6\\
			R-101 + GCP$_{D}$      & 36.5 & 58.2 & 38.9 & 17.3 & 39.9 & 53.5\\
			R-101 + GCP$_{M}$    & \textbf{36.7} & \textbf{58.7} & \textbf{39.1} & \textbf{17.6} & \textbf{39.9} & \textbf{53.7} \\
			\hline
		\end{tabular}
	\end{center}
	\caption{Instance segmentation results of various deep CNN models using Mask R-CNN on COCO val2017.}
	\label{segmentation}
\end{table}

\subsection{Robustness to Distorted Examples}
Improvement of loss Lipschitzness and gradient predictiveness brought by GCP make the networks more robust to inputs perturbed by distortions. We also note similar conclusion is stated in~\cite{DBLP:conf/icml/CisseBGDU17}. To verify this point, we conduct experiments on recently introduced IMAGENET-C and IMAGENET-P benchmarks~\cite{DBLP:conf/iclr/HendrycksD19}. Different from the works~\cite{DBLP:journals/corr/abs-1709-10207,DBLP:conf/iclr/MetzenGFB17} that study  effect of adversarial distortions as a type of worst-case analysis for network robustness, these two benchmarks are designed to evaluate robustness of deep CNNs to common image corruptions and perturbations, which have a connection with adversarial distortions and play a key role in safety-critical applications. 

The IMAGENET-C benchmark performs fifty types of corruptions (e.g., noise, blur, weather and digital) on validation set of ImageNet, and each type of corruption has five levels of severity. The IMAGENET-P benchmark generates a series of perturbation
sequences on validation set of ImageNet by performing more than ten types of perturbations, such as motion and zoom blur, brightness, translation, rotation, scale and tilt perturbations. Following the standard protocol in~\cite{DBLP:conf/iclr/HendrycksD19}, we train all CNN models on training set (clean images) of ImageNet, and report the results on IMAGENET-C and IMAGENET-P benchmarks. The evaluation metrics include mean Corruption Error (mCE) and Relative mean Corruption Errors (Relative mCE) for IMAGENET-C, mean Flip Rate (mFR) and mean Top-5 Distance (mT5D) for IMAGENET-P. For details of the metrics one can refer to~\cite{DBLP:conf/iclr/HendrycksD19}. \emph{Note that lower values  indicate better performance for all evaluation metrics.}

For a fair comparison, we employ evaluation \href{https://github.com/hendrycks/robustness}{code} released by the authors. Note that AlexNet~\cite{nips2012cnn} is a baseline model, which obtains value of 100 for all evaluation metrics. The results of different deep CNNs with GAP and GCP are given in Table~\ref{adversarial}, from which we can see that the networks with GCP significantly outperform GAP-based ones, suggesting that GCP can greatly improve the robustness of deep CNNs to common image corruptions and perturbations. Note that VGG-VD19~\cite{Simonyan15} and VGG-VD19 with BN achieve 88.9, 122.9, 66.9, 78.6 and 81.6, 111.1, 65.1, 80.5 in terms of mCE, Relative mCE, mFR, mT5D, respectively. Albeit BN improves the Lipschitzness, it is not robust to perturbations. In contrast, GCP is robust to both corruptions and perturbations.  In \cite{DBLP:conf/iclr/HendrycksD19}, many schemes are suggested to improve the robustness to corruptions and perturbations, and our work shows that GCP is a novel and promising solution. Moreover, it is potential to combine GCP with other schemes for further improvement. 

\subsection{Generalization Ability to Other Tasks}

Since GCP usually allows deep CNNs converge to better local minima, the networks with GCP pretrained on large-scale dataset can provide a better initialization model to other vision tasks. That is, they may have good generalization ability. To verify this, we first train the networks with GCP on ImageNet, and then directly apply them to object detection and instance segmentation tasks. Specifically, using ResNet-50 and ResNet-101 as backbone models, we compare performance of the networks trained with GAP and GCP on MS COCO~\cite{MSCOCO} using Faster R-CNN~\cite{DBLP:journals/pami/RenHG017} and Mask R-CNN~\cite{DBLP:conf/iccv/HeGDG17} as basic detectors. For training networks with GCP, Li et al.~\cite{LiXWZ17} suggest no down-sampling in \emph{conv5\_1}. This increases resolution of the last feature maps while resulting in larger computational cost, especially for object detection and instance segmentation where large-size input images are required. To handle this issue, we introduce two strategies: (1) we still use down-sampling as done in the original ResNet, and the method is indicated by GCP$_{D}$; (2) a max-pooling layer with a step size 2 is inserted before \emph{conv5\_1} for down-sampling, indicted by GCP$_{M}$. 

For a fair comparison, all methods are implemented by MMDetection toolkit~\cite{mmdetection} with the same (default) settings. Specifically, the shorter side of input images are resized to 800, and SGD is used to optimize the networks with a weight decay of 1e-4, a momentum of 0.9 and a mini-batch size of 8. All detectors are trained within 12 epochs on train2017 of COCO, where the learning rate is initialized to 0.01 and is decreased by a factor of 10 after 8 and 11 epochs, respectively. The results on val2017 are reported for comparison. As listed in Table~\ref{detection}, GCP$_{M}$ outperforms GCP$_{D}$ while both of them are superior to GAP. Specifically, for ResNet-50, GCP$_{M}$ improves GAP by 0.7\% in terms of AP using Faster R-CNN and Mask R-CNN as detectors. For ResNet-101, GCP$_{M}$ outperforms GAP by 0.9\% and 1.3\% for Faster R-CNN and Mask R-CNN, respectively. For results of instance segmentation in Table~\ref{segmentation}, GCP$_{M}$ improves GAP by 0.6\% and 0.8\% using ResNet-50 (R-50) and ResNet-101 (R-101) as backbone models, respectively. Note that GCP brings more improvement when backbone model and detector are stronger. These results show the pre-trained networks with GCP on large-scale datasets can well generalize to different vision tasks, indicating the networks with GCP can provide better initialization models.   

\begin{table}
	\begin{center}
		\footnotesize
		\begin{tabular}{|l|c|c|c|c|c|c|}
			\hline
			Backbone & Method  & Top-1  & Top-5 & AP & AP$_{50}$ & AP$_{75}$  \\
			\hline\hline
			\multirow{2}{*}{ResNet-18} & w/o DS & \textbf{75.47} & \textbf{92.23} & 30.0 & 50.7 & 31.4 \\
			& w/ DS  & 74.48  & 91.68 & \textbf{30.3} & \textbf{51.0} & \textbf{32.2} \\
			\hline
			\multirow{2}{*}{ResNet-50} & w/o DS & \textbf{78.56} & \textbf{94.16} & 36.6 & 58.4 & 39.5 \\
			& w/ DS  & 78.10 & 94.09  & \textbf{36.8} & \textbf{58.5} & \textbf{39.7} \\
			\hline
			\multirow{2}{*}{ResNet-101} & w/o DS & \textbf{79.48} & \textbf{94.75} & 39.5 & 60.7 & 43.1 \\
			& w/ DS & 79.11  & 94.56  & \textbf{39.6} & \textbf{60.9} & \textbf{43.4} \\
			\hline
		\end{tabular}
	\end{center}
	\caption{Comparison of GCP with or without down-sampling (DS) on ImageNet (columns 3 and 4) and MS COCO (columns 5, 6, and 7). Here, Faster R-CNN is used for object detection.}
	\label{table:lim_LFM}
\end{table}

As described above, integration of GCP into ResNets discards down-sampling (DS) operation in \emph{conv5\_1} to obtain more sampling features for more promising classification performance. However, it decreases the resolution of \emph{conv5\_x} and increases computing cost, especially for large-size input images. Here, we assess its effect on performance of GCP. Specifically, we employ ResNet-18, ResNet-50 and ResNet-101 as backbone models, and compare the results of GCP with or without DS (i.e., $conv5\_1$ with a stride of 2). For object detection, we use Faster R-CNN as the basic detector. As shown in Table~\ref{table:lim_LFM}, GCP with DS is inferior to one without DS on ImageNet classification for each model, but achieves better performance on object detection. These results suggest that DS can be introduced for balancing classification accuracy, performance of object detection and model complexity for the networks trained with GCP. 


\section{Conclusion}

In this paper, we made an attempt to analyze the effectiveness of GCP on deep CNNs from an optimization perspective. Specifically, we showed that GCP has the ability to improve Lipschitzness of loss and predictiveness of gradient in context of deep CNNs, and discussed the connection between GCP and second-order optimization. Our findings can account for several merits of GCP for training deep CNNs that have not been recognized previously or fully explored, including significant acceleration of network convergence, stronger robustness to distorted examples, and good generalization ability to different vision tasks. The extensive experimental results provide strong support to our findings. Our work provides an inspiring view to understand GCP and may help researchers explore more merits of GCP in context of deep CNNs. In future, we will investigate the theoretical proofs on smoothing effect of GCP and the rigorous connection to second-order optimization.

{\small
	\bibliographystyle{ieee_fullname}
	\bibliography{arxiv_4421}
}

\section*{Appendix I: Implementation Details for Analyzing Smoothing Effect of GCP}

In Section 3.2, we analyze smoothing effect of GCP on deep CNNs in terms of the Lipschitzness of optimization loss and the predictiveness of gradients. Specifically, the Lipschitzness of optimization loss is measured by 
\begin{align}\label{equ1}
\vartriangle_{l} = \mathcal{L}(\mathbf{X} + \eta_{l}\nabla\mathcal{L}(\mathbf{X})), \eta_{l} \in [a, b],
\end{align}
and the predictiveness of gradients is measured by 
\begin{align}\label{equ2}
\vartriangle_{g} = \|\nabla\mathcal{L}(\mathbf{X})- \nabla\mathcal{L}(\mathbf{X}+\eta_{g}\nabla\mathcal{L}(\mathbf{X}))\|_{2}, \eta_{g} \in [a, b],
\end{align}
where $\mathbf{X}$ is the input; $ \nabla\mathcal{L}(\mathbf{X}) $ indicates the gradient of loss with respect to the input $\mathbf{X}$; $\eta_{l}$ and $\eta_{g}$ indicate step sizes of gradient descent. 

To assess effect of GCP on the whole CNN models following~\cite{DBLP:conf/nips/SanturkarTIM18}, we employ output of the first convolution layer as $\mathbf{X}$ to compute Eqns.~(\ref{equ1}) and~(\ref{equ2}). Note that the experiments in Section 4.2 demonstrate that the networks with GCP is more robust to input images with perturbations, comparing with those based on GAP. Accordingly, optimization loss of the networks with GCP also is more stable to input images with perturbations. For clear illustration, we calculate the ranges of $\vartriangle_{l}$ and $\vartriangle_{g}$ every 1,000 and 500 training steps for MobileNetV2 and ResNet-18, respectively. For calculating the ranges of $\vartriangle_{l}$ and $\vartriangle_{g}$, we uniformly sample 50 points of $\eta_{l}$ (and $\eta_{g}$) from $[0.045,\,1.5]$ and $[0.1,\,75]$ for MobileNetV2 and ResNet-18, respectively. Then, we plot the ranges of $\vartriangle_{l}$ and $\vartriangle_{g}$ determined by the minimum and maximum of the 50 sampled points.


\section*{Appendix II: Derivations of Eqn. (6) and Eqn. (7)}

As described in Section 3.3, the gradient of the loss with respect to the input $\mathbf{X}$ through GCP layer can be calculated as
\begin{align}\label{equ3}
\frac{\partial \mathcal{L}}{\partial \mathbf{X}} = & 2\mathbf{JX}\bigg[\mathbf{U}\bigg(\bigg(\mathbf{K}^{T} \circ \bigg(\mathbf{U}^{T} 2 \bigg(\frac{\partial \mathcal{L}}{\partial \mathbf{Z}_{\text{GCP}}} \bigg)_{\text{sym}} \mathbf{U}\boldsymbol{\Lambda}^{\frac{1}{2}}\bigg)\bigg)+ \nonumber \\
&\bigg(\frac{1}{2}\boldsymbol{\Lambda}^{-\frac{1}{2}}\mathbf{U}^{T}
\frac{\partial \mathcal{L}}{\partial \mathbf{Z}_{\text{GCP}}}\mathbf{U}\bigg)_{\text{diag}}\bigg)\mathbf{U}^{T}\bigg]_{\text{sym}}.
\end{align}
Here, we give detailed derivations of Eqn.~(\ref{equ3}) as follows. To perform GCP, we compute the square root of sample covariance matrix of features  $\mathbf{X}\in \mathbb{R}^{N\times D}$ as
\begin{align}\label{equ4}
\mathbf{Z}_{\text{GCP}}=\boldsymbol\Sigma^{\frac{1}{2}}=(\mathbf{X}^{T}\mathbf{JX})^{\frac{1}{2}} = \mathbf{U} \boldsymbol\Lambda^{\frac{1}{2}} \mathbf{U}^{T}, 
\end{align}
where $\mathbf{U}$ and $\boldsymbol{\Lambda}$ are the matrix of eigenvectors and the diagonal matrix of eigenvalues of sample covariance $\boldsymbol\Sigma$, respectively. As shown in~\cite{LiXWZ17}, $\frac{\partial \mathcal{L}}{\partial \mathbf{X}}$ can be calculated as
\begin{align}\label{equ5}
\frac{\partial \mathcal{L}}{\partial \mathbf{X}}=
\mathbf{JX}\bigg(\frac{\partial \mathcal{L}}{\partial\boldsymbol\Sigma}+\bigg(\frac{\partial \mathcal{L}}{\partial \boldsymbol\Sigma}\bigg)^{T}\bigg),
\end{align}
\begin{align}\label{equ6}
\frac{\partial \mathcal{L}}{\partial \boldsymbol\Sigma}=
\mathbf{U}\bigg(\bigg(\mathbf{K}^{T} \circ \bigg(\mathbf{U}^{T}\frac{\partial \mathcal{L}}{\partial \mathbf{U}}\bigg)\bigg)+\bigg(\frac{\partial \mathcal{L}}{\partial \boldsymbol\Lambda}\bigg)_{\text{diag}}\bigg)\mathbf{U}^{T},
\end{align}
\begin{align}\label{equ7}
\frac{\partial \mathcal{L}}{\partial \mathbf{U}} = \bigg(\frac{\partial \mathcal{L}}{\partial \mathbf{Z}_{\text{GCP}}}+\bigg(\frac{\partial \mathcal{L}}{\partial \mathbf{Z}_{\text{GCP}}}\bigg)^{T}\bigg)\mathbf{U}\boldsymbol\Lambda^{\frac{1}{2}},
\end{align}
\begin{align}\label{equ8}
\frac{\partial \mathcal{L}}{\partial \boldsymbol\Lambda} = \frac{1}{2}\bigg(\boldsymbol\Lambda^{-\frac{1}{2}}\mathbf{U}^{T}\frac{\partial \mathcal{L}}{\partial \mathbf{Z}_{\text{GCP}}} \mathbf{U}\bigg)_{\text{diag}}.
\end{align}
Let $(\mathbf{A})_{\text{sym}} = \frac{1}{2}(\mathbf{A}+\mathbf{A}^{T})$, we can rewrite  Eqn.~(\ref{equ5}) and Eqn.~(\ref{equ7}) as
\begin{align}\label{equ9}
\frac{\partial \mathcal{L}}{\partial \mathbf{X}}=2\mathbf{JX}\bigg(\frac{\partial \mathcal{L}}{\partial \boldsymbol\Sigma}\bigg)_{\text{sym}},
\end{align}
\begin{align}\label{equ10}
\frac{\partial \mathcal{L}}{\partial \mathbf{U}}=2\bigg(\frac{\partial \mathcal{L}}{\partial \mathbf{Z}_{\text{GCP}}}\bigg)_{\text{sym}}\mathbf{U}\boldsymbol\Lambda^{\frac{1}{2}}.
\end{align}
By substituting Eqns.~(\ref{equ6}), ~(\ref{equ8}) and ~(\ref{equ10}) into Eqn.~(\ref{equ9}), we achieve
\begin{align}\label{equ11}
\frac{\partial \mathcal{L}}{\partial \mathbf{X}}
&=2\mathbf{JX}\bigg(\frac{\partial \mathcal{L}}{\partial \boldsymbol\Sigma}\bigg)_{\text{sym}} \nonumber \\ 
&=2\mathbf{JX}\bigg(\mathbf{U}\bigg(\bigg(\mathbf{K}^{T} \circ \bigg(\mathbf{U}^{T}\frac{\partial \mathcal{L}}{\partial \mathbf{U}}\bigg)\bigg)+\bigg(\frac{\partial \mathcal{L}}{\partial \boldsymbol\Lambda}\bigg)_{\text{diag}}\bigg)\mathbf{U}^{T}\bigg)_{\text{sym}}  \\ 
&=2\mathbf{JX}\bigg[\mathbf{U}\bigg(\bigg(\mathbf{K}^{T} \circ \bigg(\mathbf{U}^{T}2\bigg(\frac{\partial \mathcal{L}}{\partial \mathbf{Z}_{\text{GCP}}}\bigg)_{\text{sym}}\mathbf{U}\boldsymbol\Lambda^{\frac{1}{2}}\bigg)\bigg)\nonumber \\
&+\bigg(\frac{1}{2}\boldsymbol\Lambda^{-\frac{1}{2}}\mathbf{U}^{T}\frac{\partial \mathcal{L}}{\partial \mathbf{Z}_{\text{GCP}}} \mathbf{U}\bigg)_{\text{diag}}\bigg)\mathbf{U}^{T}\bigg]_{\text{sym}}. \nonumber
\end{align} 
So far, we obtain Eqn.~(\ref{equ3}). 

With some assumptions and simplification, Eqn.~(\ref{equ3}) can be trimmed as
\begin{align}\label{equ12}
\frac{\partial \mathcal{L}}{\partial \mathbf{X}} \thickapprox 2\mathbf{JX}\bigg(2\mathbf{K}^{T} \circ \boldsymbol\Lambda^{\frac{1}{2}}+\frac{1}{2}\boldsymbol\Lambda^{-\frac{1}{2}}\bigg)\frac{\partial \mathcal{L}}{\partial \mathbf{Z}_{\text{GCP}}},
\end{align}
where $\circ$ denotes matrix Hadamard product. In the following, we explain how we obtain Eqn.~(\ref{equ12}). Specifically, we simplify Eqn.~(\ref{equ3}) by neglecting $(\centerdot)_{\text{sym}}$ and $(\centerdot)_{\text{diag}}$ operations. Thus, Eqn.~(\ref{equ3}) can be approximated by
\begin{align}\label{equ13}
\frac{\partial \mathcal{L}}{\partial \mathbf{X}} \thickapprox  & 2\mathbf{JX}\bigg[\mathbf{U}\bigg(\mathbf{K}^{T} \circ \bigg(\mathbf{U}^{T} 2 \frac{\partial \mathcal{L}}{\partial \mathbf{Z}_{\text{GCP}}}  \mathbf{U}\boldsymbol\Lambda^{\frac{1}{2}}\bigg)+ \nonumber \\
 &\frac{1}{2}\boldsymbol\Lambda^{-\frac{1}{2}}\mathbf{U}^{T}
\frac{\partial \mathcal{L}}{\partial \mathbf{Z}_{\text{GCP}}}\mathbf{U}\bigg)\mathbf{U}^{T}\bigg].
\end{align}
Then, we assume that matrix multiplications between diagonal matrix $\boldsymbol\Lambda$ and orthogonal matrix $\mathbf{U}$ (or symmetric matrix $\frac{\partial \mathcal{L}}{\partial \mathbf{Z}_{\text{GCP}}}$) in Eqn.~(\ref{equ13}) satisfy the commutative law of multiplication. So Eqn.~(\ref{equ13}) can be trimmed as
\begin{align}\label{equ14}
\frac{\partial \mathcal{L}}{\partial \mathbf{X}} \thickapprox & 2\mathbf{JX}\bigg[\mathbf{U}\bigg(\mathbf{K}^{T} \circ \bigg(\mathbf{U}^{T} 2\boldsymbol\Lambda^{\frac{1}{2} }\frac{\partial \mathcal{L}}{\partial \mathbf{Z}_{\text{GCP}}}  \mathbf{U}\bigg)+ \nonumber \\
 &\frac{1}{2}\mathbf{U}^{T}\boldsymbol\Lambda^{-\frac{1}{2}}
\frac{\partial \mathcal{L}}{\partial \mathbf{Z}_{\text{GCP}}}\mathbf{U}\bigg)\mathbf{U}^{T}\bigg].
\end{align}
Finally, we assume that the mask matrix $\mathbf{K}$ only has effect on the diagonal matrix of eigenvalues $\boldsymbol\Lambda$. So we have
\begin{align}\label{equ15}
\frac{\partial \mathcal{L}}{\partial \mathbf{X}} \thickapprox & 2\mathbf{JX}\bigg[\mathbf{U}\bigg(\mathbf{U}^{T} 2\mathbf{K}^{T} \circ \boldsymbol\Lambda^{\frac{1}{2} }\frac{\partial \mathcal{L}}{\partial \mathbf{Z}_{\text{GCP}}}  \mathbf{U}+ \nonumber \\
& \frac{1}{2}\mathbf{U}^{T}\boldsymbol\Lambda^{-\frac{1}{2}}
\frac{\partial \mathcal{L}}{\partial \mathbf{Z}_{\text{GCP}}}\mathbf{U}\bigg)\mathbf{U}^{T}\bigg]\nonumber\\
= &
2\mathbf{JX}\bigg[\mathbf{U}\bigg(\mathbf{U}^{T}\bigg(2\mathbf{K}^{T} \circ \boldsymbol\Lambda^{\frac{1}{2} }\frac{\partial \mathcal{L}}{\partial \mathbf{Z}_{\text{GCP}}} + \nonumber \\
& \frac{1}{2}\boldsymbol\Lambda^{-\frac{1}{2}} \frac{\partial \mathcal{L}}{\partial \mathbf{Z}_{\text{GCP}}}\bigg)\mathbf{U}\bigg)\mathbf{U}^{T}\bigg] \\
= &
2\mathbf{JX}\bigg(2\mathbf{K}^{T} \circ \boldsymbol\Lambda^{\frac{1}{2} } +
\frac{1}{2}\boldsymbol\Lambda^{-\frac{1}{2}}\bigg)
\frac{\partial \mathcal{L}}{\partial \mathbf{Z}_{\text{GCP}}}.\nonumber
\end{align} 
Note that, in practice, Eqn.~(\ref{equ15}) is not employed for back-propagation of GCP, but provides a simplified form of Eqn.~(\ref{equ3}) for discussion on connection with second-order optimization in context of deep CNNs.

\section*{Appendix III: Convergence Curves of Networks with GCP under Various Dimensions of Input}

\begin{figure}[t]
	\begin{center}
		\includegraphics[width=1.0\linewidth]{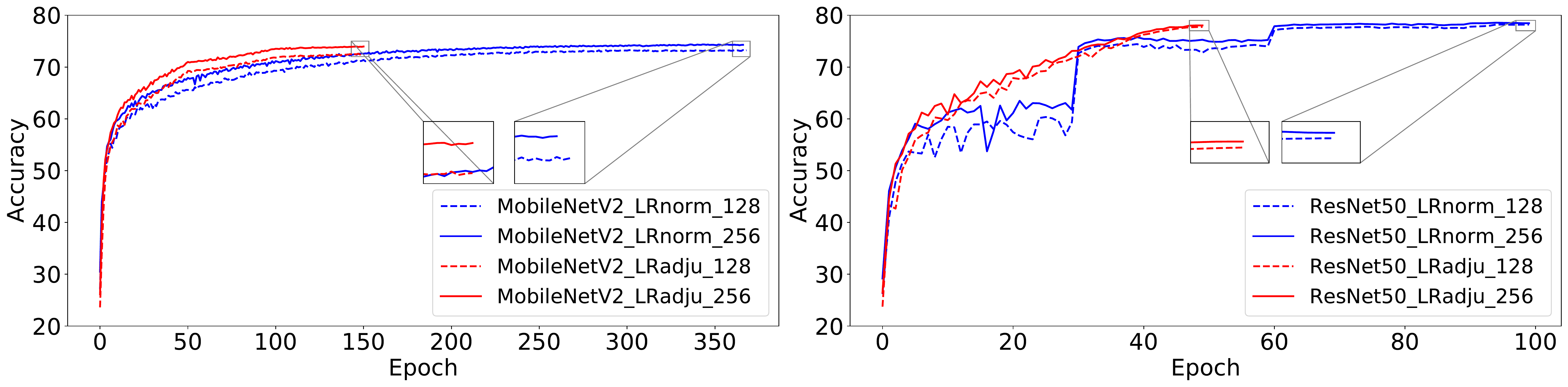}
	\end{center}
	\caption{Convergence curves of MobileNetV2 and ResNet-50 trained with GCP under different settings of $lr$ and various dimension ($D$) of input features on ImageNet.}
	\label{fig:cc}
\end{figure}

In Table 5 of Section 4.1,  we gave the results of MobileNetV2 and ResNet-50 with GCP under different settings of $lr$ (i.e., $LR_{norm}$ and $LR_{adju}$) and various dimension (i.e., $D=256$ and $D=128$)  of input features on ImageNet. Figure~\ref{fig:cc} illustrates their corresponding convergence curves, from which we can see that lower-dimensional covariance representations (COV-Reps) share similar behavior with higher-dimensional COV-Reps, but the lower-dimensional COV-Reps suffer from larger performance degradation in the case of faster convergence (i.e., $LR_{adju}$). 

\section*{Appendix IV: Implementation Details on Applying Pre-trained Networks with GCP to Other Vision Tasks}

To apply the pre-trained networks with GCP to object detection and instance segmentation on MS COCO, we adopt the same strategy with the original GAP-based CNN models~\cite{DBLP:conf/iccv/HeGDG17,DBLP:journals/pami/RenHG017} and make a modification, i.e., increasing resolution of feature maps in the last stage. The detailed steps are described as follows. All detectors are implemented using MMDetection toolkit~\cite{mmdetection}.
\begin{itemize}
	\item[S\_I:]Pre-training the networks with GCP on ImageNet~\cite{imagenet_cvpr09} without down-sampling in \emph{conv5\_1} as suggested in~\cite{LiXWZ17};
	\item[S\_II:]Discarding the GCP layer and the classifier, while introducing Region Proposal Networks (RPN)~\cite{DBLP:journals/pami/RenHG017} and  Region of Interest (ROI) Pooling~\cite{DBLP:conf/iccv/HeGDG17,DBLP:journals/pami/RenHG017};
	\item[S\_III:]Increasing resolution of feature maps in the last stage using GCP$_D$ (i.e., use of down-sampling as done in the original ResNet) and GCP$_M$ (i.e., a max-pooling layer with a step size 2 is inserted before \emph{conv5\_1}) strategies, while introducing feature pyramid networks (FPN)~\cite{DBLP:conf/cvpr/LinDGHHB17}; 
	\item[S\_IV:] Fine-tuning the whole networks in S\_III on MS COCO~\cite{MSCOCO} using the same hyper-parameters with those of the original GAP-based CNN models.
\end{itemize}

\begin{table*}[t]
	\centering
	\renewcommand\arraystretch{1.4}
	\caption{Comparison of GCP and GAP using various ResNets in terms of network parameters, floating point operations per second (FLOPs), training or inference time per image, and classification accuracy.}\smallskip
	\begin{tabular}{lcccccc}
		\hline
		Methods & Parameter & GFLOPs. & Training time (ms) & Inference time (ms)  &Top-1 Err. (\%) & Top-5 Err. (\%)\\
		\hline
		ResNet18 + GAP & 11.69M & 1.81 & 0.77  & 0.60  & 70.47 & 89.59 \\
		ResNet18 + GCP & 19.60M & 3.11 & 1.21  & 0.85  & 75.07 & 92.14 \\  
		\hline
		ResNet34 + GAP & 21.80M & 3.66 & 1.17  & 0.88  & 74.19 & 91.60 \\
		ResNet34 + GCP & 29.71M & 5.56 & 1.61   &1.10  &  76.80 & 93.11 \\ 
		\hline
		ResNet50 + GAP & 25.56M & 3.86 & 1.85  &1.29 & 76.02 & 92.97 \\
		ResNet50 + GCP & 32.32M & 6.19 & 2.22  &1.49 & 78.56 & 93.72 \\ 
		\hline
		ResNet101 + GAP & 44.55M & 7.57 & 2.79 &1.72 & 77.67 & 93.83 \\
		ResNet101 + GCP & 51.31M & 9.90 &3.14 &1.83  & 79.47 & 94.30 \\ 
		\hline
		ResNet152 + GAP & 60.19M & 11.28 & 3.54 &2.55 & 78.13 & 94.04 \\
		\hline
	\end{tabular}
	\label{table:comparsion}	
\end{table*}

\section*{Appendix V: Computational Comparison of GCP and GAP}

Here, we compare GCP and GAP in terms of computational cost. The experiments are conducted on large-scale ImageNet using ResNet-18, ResNet-34, ResNet-50 and ResNet-101 as backbone models. The evaluation metrics include network parameters, floating point operations per second (FLOPs), training or inference time per image, and Top-1/Top-5 accuracies. For GCP, size of covariance representations is set to 8k. All models are trained with the same experimental settings and run on a workstation equipped with four Titan Xp GPUs, two Intel(R) Xeon Silver 4112 CPUs @ 2.60GHz, 64G RAM and 480 GB INTEL SSD.  From the results in Table~\ref{table:comparsion}, we can see that GCP introduces extra $\sim$7M parameters, $\sim$0.4ms training time and $\sim$0.2ms inference time, but increase about 4.6\%, 2.6\%, 1.5\% and 1.8\% Top-1 accuracies over GAP-based ResNet-18, ResNet-34, ResNet-50 and ResNet-101, respectively. Besides, GCP achieves matching performance using  much lower computational complexity than GAP (e.g., ResNet34+GCP \emph{vs.} ResNet101+GAP and ResNet50+GCP \emph{vs.} ResNet152+GAP). Additionally, GCP with similar computational complexity achieves much better performance than GAP (e.g., ResNet18+GCP \emph{vs.} ResNet34+GAP and ResNet50+GCP \emph{vs.} ResNet101+GAP). Note that we discard down-sampling operation in $conv5\_x$ for GCP with ResNets, which significantly increases FLOPs. When we use this down-sampling operation, GCP shares similar FLOPs with GAP, leading slight performance decrease. 

\end{document}